\documentclass[lettersize,journal]{IEEEtran}
\usepackage{amsmath,amsfonts}
\usepackage{algorithmic}
\usepackage{algorithm}
\usepackage{array}
\usepackage[caption=false,font=normalsize,labelfont=sf,textfont=sf]{subfig}
\usepackage{textcomp}
\usepackage{enumitem}
\usepackage{stfloats}
\usepackage{url}
\usepackage{verbatim}
\usepackage{graphicx}
\usepackage{cite}
\usepackage{multirow}
\usepackage{booktabs} 
\def\BibTeX{{\rm B\kern-.05em{\sc i\kern-.025em b}\kern-.08em
    T\kern-.1667em\lower.7ex\hbox{E}\kern-.125emX}}
\usepackage{balance}
\usepackage[colorlinks,bookmarks=false,citecolor=green]{hyperref}
\usepackage{tikz}
\usepackage{pgfplots}
\pgfplotsset{compat=1.17}

\begin{document}

\title{Lens Privacy Sealing: A New Benchmark and Method \\ for Physical Privacy-Preserving Action Recognition}%

\author{
  Mengyuan Liu, 
  Ziyi Wang$^{\dagger}$, 
  Peiming Li$^{\dagger}$,  
  Junsong Yuan

  \thanks{
This work was supported by National Natural Science Foundation of China (No. 62473007), Guangdong Outstanding Youth Fund (No. 2026B1515020015), Shenzhen Innovation in Science and Technology Foundation for The Excellent Youth Scholars (No. RCYX20231211090248064). \protect\\
M. Liu, Z. Wang, and P. Li are with the State Key Laboratory of General Artificial Intelligence, Peking University, Shenzhen Graduate School ($^{\dagger}$Corresponding author. Email: \{ziyiwang, lipeiming1001\}@stu.pku.edu.cn). \protect\\
J. Yuan is with the Department of Computer Science and Engineering, State University of New York at Buffalo.}
}

\markboth{IEEE TRANSACTIONS ON IMAGE PROCESSING}%
{Liu \MakeLowercase{\textit{et al.}}: Lens Privacy Sealing: A New Benchmark and Method for Physical Privacy-Preserving Action Recognition}

\maketitle

\begin{abstract}
RGB camera-based surveillance systems enable human action recognition for public safety and healthcare, yet raise serious privacy concerns. Existing methods rely on post-capture algorithms, which fail to protect privacy during data acquisition. We propose Lens Privacy Sealing (LPS), a simple hardware solution that physically obscures camera lenses with adjustable laminating film, providing pre-sensor privacy protection at minimal cost. Unlike software methods or expensive engineered optics, LPS achieves strong privacy through stochastic multi-layer scattering that is physically irreversible. We introduce the P$^3$AR dataset for privacy-preserving action recognition, featuring both large-scale replay-captured (P$^3$AR-NTU, 114K videos) and real-world collected (P$^3$AR-PKU) subsets with privacy attribute annotations. To handle video degradation from LPS, we propose MSPNet, a single-stage framework incorporating Inter-Frame Noise Suppressor (IFNS) and Cross-Frame Semantic Aggregator (CFSA), enhanced by contrastive language-image pre-training for robust semantic extraction. Extensive experiments demonstrate that MSPNet with IFNS and CFSA nearly doubles action recognition accuracy compared to baseline methods while suppressing identity recognition to low levels. Comprehensive validation shows LPS achieves a superior privacy-utility trade-off compared to state-of-the-art hardware methods, resists reconstruction attacks including PSF inversion and data-driven recovery, and generalizes robustly across optical configurations and challenging environments. Code is available at https://github.com/wangzy01/MSPNet.
\end{abstract}

\begin{IEEEkeywords}
action recognition, privacy protection
\end{IEEEkeywords}

\section{Introduction}
\IEEEPARstart{T}{he} widespread deployment of RGB camera-based surveillance has enabled Human Action Recognition (HAR) in applications such as public safety~\cite{10487798}, healthcare supervision~\cite{Gu2021ASO,pu2024learning}, smart homes~\cite{myung2024degcn,tang2024learnable}, and human-robot interaction~\cite{wang2025recognizing}. However, these systems often collect vast amounts of personal information without explicit consent. Despite substantial advancements in HAR, privacy protection remains a critical concern. This motivates the development of action recognition methods that enhance privacy, which not only increases public acceptance but also embodies responsible integration of technology and ethics.

Existing privacy-preserving methods primarily rely on post-capture algorithms such as pixelation~\cite{Maaten2008DimensionalityRA}, blurring~\cite{PADILLALOPEZ20154177}, face replacement~\cite{Chen2007ToolsFP}, low resolution~\cite{Ryoo2017}, and de-identification~\cite{DeepPrivacy}. Recently, Ren \textit{et al.}~\cite{Ren2018LearningTA} use adversarial training to anonymize faces prior to action detection. However, other personal information, such as clothing, continues to pose privacy risks, and these methods still require the use of original high-resolution videos, which inherently increases privacy vulnerabilities. As for human action recognition tasks, Ryoo \textit{et al.}\cite{Ryoo2017ExtremeLR, Ryoo2017} use low-resolution cameras to anonymize videos while maintaining action recognition capability. Besides, Pittaluga \textit{et al.}\cite{7298628} explore the use of defocused lenses to achieve privacy in workspaces. 
Generally, these methods often fail to adequately mitigate privacy risks during data capture, leaving devices vulnerable to unauthorized access or tampering.

\begin{center}
\begin{figure}[t]
\begin{center}
\includegraphics[width=0.998\columnwidth]{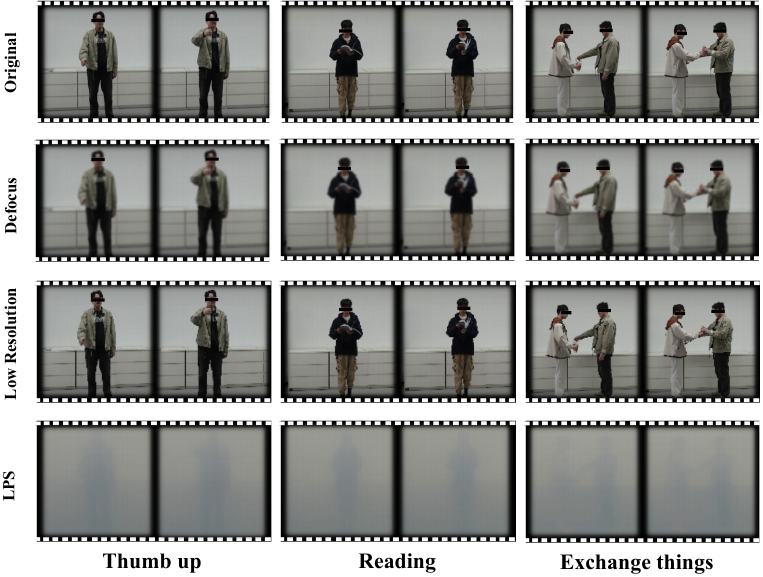}  
\end{center}
\vspace{-4mm}
\caption{\textbf{Comparison with Existing Privacy-Preserving Methods}. Existing privacy-preserving methods predominantly rely on post-capture algorithms applied to original RGB videos, such as defocusing and low-resolution processing. However, these methods often fall short of fully mitigating privacy risk during data capture, as the original RGB videos still exist. In contrast, our proposed Lens Privacy Sealing (LPS) offers a physical solution that eliminates the need for generating original RGB videos, achieving privacy protection with minimal hardware modifications.}
\label{fig:LPS-degraded}   
\vspace{-1em}
\end{figure}
\end{center}

To address this gap, we introduce a physical privacy-preserving solution called Lens Privacy Sealing (LPS), which involves physically covering RGB camera lenses with laminating film, with adjustable layers for users to manually control the level of privacy protection. Furthermore, due to the wide availability and low cost of laminated film materials, LPS can be easily integrated with existing devices. Fig.~\ref{fig:LPS-degraded} presents a qualitative comparison between LPS and existing privacy-preserving methods. Defocusing and low-resolution techniques have been shown to provide suboptimal privacy protection effectiveness~\cite{9207852}. In comparison, LPS not only obscures faces but also hinders the recognition of other privacy-sensitive attributes, such as age, skin color, clothing, and gender.

To evaluate our approach, we collect a P$^3$AR dataset for Physical Privacy-Preserving Action Recognition. P$^3$AR employs two construction methods: recapturing existing datasets (P$^3$AR-NTU) and collecting real-world data (P$^3$AR-PKU) with 8--10 layers of laminating film from multiple viewpoints.

Conventional HAR approaches typically follow a two-stage process~\cite{8955791,rcnn}, where detecting human subjects from degraded, low-quality videos is often challenging, leading to suboptimal performance. To overcome the unique challenge of video degradation, we propose a single-stage Motion Semantic Prompting Network (MSPNet). MSPNet is a single-stage framework tailored for degraded videos, balancing privacy protection and recognition accuracy. Its key component, the Inter-Frame Noise Suppressor (IFNS), is inspired by video residuals for pose estimation~\cite{Liu2021DeepDC}. IFNS employs inter-frame subtraction to suppress noise from physical privacy measures while preserving motion contours. Additionally, the Cross-Frame Semantic Aggregator (CFSA) combines different frame differences using dynamic weighting to capture richer spatio-temporal features. In practice, surveillance devices are typically fixed, and the background elements in the video are relatively static. Therefore, this integrated approach eliminates the human detection step commonly found in traditional two-stage processes, enabling the direct recognition of human movements in degraded video. This approach helps enhance privacy protection during the data acquisition phase. Inspired by existing advances in contrastive language-image pre-training~\cite{Radford2021LearningTV, Narasimhan2021CLIPItLV, Ni2022ExpandingLP, Sain2023CLIPFA, Xu2021VideoCLIPCP,wang2026unimotion}, we leverage text as a supervisory signal to enhance the semantic richness of our model. We use CLIP~\cite{Radford2021LearningTV} to adapt its generalization ability from images to degraded videos by integrating spatio-temporal information through an Inter-frame Interaction Transformer and a Spatio-temporal Fusion Transformer.

We conduct extensive experiments on both the P$^3$AR-NTU and P$^3$AR-PKU for human action recognition. Our MSPNet increases the action recognition accuracy on P$^3$AR-NTU from 38.38\% to 67.90\%. We achieve a Subject ID Accuracy ($ACC_S$) below 10\%, and in real-world scenarios, we maintain an action recognition accuracy of 74.83\% while keeping $ACC_S$ similarly low. Our assessment and visualization show that LPS effectively balances privacy protection with high HAR accuracy.

In this paper, \textit{benchmarking physical privacy-preserving action recognition} refers to systematically evaluating both utility (action recognition accuracy) and privacy leakage ($ACC_S$) across diverse physical and algorithmic privacy mechanisms under a unified protocol. This benchmarking focuses on representative physical degradations for action recognition, rather than exhaustively covering all privacy-preserving imaging tasks. In summary, our main contributions are as follows.
\begin{itemize}[leftmargin=*]
\item Unlike algorithmic methods, we propose Lens Privacy Sealing (LPS), a simple, affordable, hardware-based solution to improve privacy protection during video capture. By applying adjustable laminating films to camera lenses, LPS provides customizable privacy levels, offering a cost-effective approach to physical privacy protection in RGB camera-based surveillance systems.
\item A new P$^3$AR dataset has been collected, contributing to the development and testing of physical privacy-preserving action recognition. P$^3$AR dataset presents unique challenges due to video degradation from privacy measures and serves as a reliable benchmark for future research.
\item To recognize actions in degraded videos caused by the LPS intervention, we propose a Motion Semantic Prompting Network (MSPNet), which integrates two key modules: the Inter-Frame Noise Suppressor (IFNS) to reduce noise from privacy measures and the Cross-Frame Semantic Aggregator (CFSA) to enhance spatio-temporal feature extraction. MSPNet boosts action recognition accuracy on P$^3$AR-NTU from 38.38\% to 67.90\%, while keeping $ACC_S$ below 10\%, surpassing state-of-the-art methods.
\item We establish a comprehensive benchmark by comparing LPS with state-of-the-art hardware methods (Coded Aperture, DiffuserCam, DyPP) and software baselines under a unified evaluation protocol, demonstrating superior privacy-utility trade-offs. Extensive validation shows MSPNet resists reconstruction attacks and generalizes across optical configurations and challenging environments.
\end{itemize}

Our target scenarios are fixed-camera deployments (surveillance, smart homes); camera motion remains a limitation that we discuss in Section~\ref{sec:robustness}.

The remainder is organized as follows. Section~\ref{sec:related} reviews related work. Section~\ref{sec:dataset} introduces Lens Privacy Sealing (LPS) and the P$^3$AR dataset. Section~\ref{sec:method} presents our Motion Semantic Prompting Network (MSPNet). Section~\ref{sec:experiment} reports experimental results and analysis. Finally, Section~\ref{sec:conclusion} concludes this work.

\renewcommand{\arraystretch}{1.1}
\begin{table*}[htbp]
\centering
\caption{Comparison of the proposed P$^3$AR dataset and some of the other publicly available datasets for human action recognition. Our dataset provides privacy attributes for privacy-preserving action recognition.}
\vspace{-2mm}
\begin{tabular}{lcccccccc}
\toprule
\textbf{Datasets} && \textbf{\#Videos} & \textbf{\#Classes} & \textbf{\#Subjects} & \textbf{\#Views} & \textbf{\#Privacy Attributes} & \textbf{Source} & \textbf{Year} \\ 
\hline
MSR-Action3D &~\cite{cvpr10msraction} & 567 & 20 & 10 & 1 & N/A & CVPR & 2010 \\ 
\hline
MSRDailyActivity3D&~\cite{cvpr12wang2012mining} & 320 & 16 & 10 & 1 & N/A & CVPR & 2012 \\ 
\hline
Multiview 3D Event&~\cite{iccv13wei2013modeling} & 3,815 & 8 & 8 & 3 & N/A & ICCV & 2013 \\ 
\hline
UWA3D Multiview&~\cite{eccv14rahmani2014hopc} & $\sim$900 & 30 & 10 & 4 & N/A & ECCV & 2014 \\ 
\hline
UWA3D Multiview II&~\cite{tpami16rahmani2016histogram} & 1,075 & 30 & 10 & 5 & N/A & TPAMI & 2016 \\ 
\hline
NTU RGB+D 60&~\cite{NTU} & 56,880 & 60 & 40 & 80 & N/A & CVPR & 2016 \\ 
\hline
SYSU 3DHOI&~\cite{tpami17hu2017jointly} & 480 & 12 & 40 & 1 & N/A & TPAMI & 2017 \\ 
\hline
NTU RGB+D 120&~\cite{tpami19liu2019ntu} & 114,480 & 120 & 106 & 155 & N/A & TPAMI & 2019 \\ 
\hline
UAV-Human&~\cite{cvpr21li2021uav} & 67,428 & 155 & 119 & - & N/A & CVPR & 2021 \\ 
\hline
HOMAGE&~\cite{cvpr21rai2021home} & 1,752 & 75 & 27 & $2\sim5$ & N/A & CVPR & 2021 \\ 
\hline
PA-HMDB51&~\cite{9207852} & 6,766 & 51 & - & - & 6 & TPAMI & 2022 \\
\hline
SMG&~\cite{ijcv23chen2023smg} & 3,712 & 16 & 40 & 1 & N/A & IJCV & 2023 \\ 
\hline
\rule{0pt}{3mm}\textbf{P$^3$AR-NTU} && \textbf{114,480} & \textbf{120} & \textbf{106} & \textbf{155} & \textbf{1} & - & - \\ 
\rule{0pt}{3mm}\textbf{P$^3$AR-PKU} && \textbf{4,500} & \textbf{60} & \textbf{25} & \textbf{3} & \textbf{6} & - & - \\ 
\bottomrule
\end{tabular}
\label{tab:dataset1}
\vspace{-1em}
\end{table*}
\renewcommand{\arraystretch}{1}

\section{Related Work}
\label{sec:related}

\subsection{Action Recognition Methods}
Previous Human Action Recognition (HAR) methods \cite{10517892,8360391,10122859,wang2026universal} commonly combine skeletal and RGB modalities to enhance performance. Hou et al. \cite{hou2016skeleton} and Wang et al. \cite{wang2016action} propose using skeleton optical spectra and joint trajectory maps, respectively, to encode the spatio-temporal information of skeleton sequences into color texture images, which are then used for behavior recognition via CNNs. However, estimating human poses in degraded, low-quality videos is nearly impossible, which is why MSPNet relies solely on the RGB modality. Early RGB-based action recognition methods mainly depend on handcrafted features, such as spatio-temporal volume approaches~\cite{gorelick2007actions} and trajectory-based methods~\cite{wang2013action}. Modern deep learning techniques can be broadly categorized into three types: two-stream 2D Convolutional Neural Networks (2D CNN)\cite{simonyan2014two}, 3D Convolutional Neural Networks (3D CNN)\cite{tran2015learning}, and Transformer-based methods~\cite{girdhar2019video}. The self-attention mechanism enables Transformers to capture long-range spatio-temporal dependencies, improving complex action recognition. This advantage has led to their widespread adoption in recent works. For example, TimeSformer~\cite{timesformer} replaces traditional convolutional operations with a divided attention design, effectively supporting long-range video modeling. UniFormer~\cite{li2022uniformer} addresses the core trade-off between local redundancy and global dependency in video understanding by seamlessly integrating 3D convolution and spatiotemporal self-attention within a unified Transformer architecture. However, these methods are mainly designed for clear RGB videos and struggle to handle the challenges posed by nonlinear optical distortions in P$^3$AR. Additionally, conventional HAR approaches typically follow a two-stage paradigm: first detecting human subjects using object detectors, then recognizing actions within the detected regions~\cite{8955791}. While effective for clear videos, this pipeline becomes problematic when image quality degrades significantly, as detection failures introduce cascading errors. MSPNet adopts a single-stage framework that directly processes full video frames, combined with motion residual-driven approach and multi-scale spatio-temporal fusion, demonstrating advantages in processing low-quality videos.

\subsection{Privacy-Preserving Methods}
Existing privacy-preserving methods can be broadly categorized into software-based post-processing and hardware-based physical protection.

\textbf{Software-based Methods.}
Previous works rely on post-capture algorithmic strategies, such as pixelation \cite{Maaten2008DimensionalityRA}, blurring \cite{PADILLALOPEZ20154177}, and lowering resolution, to reduce the exposure of sensitive information. More recent studies explore privacy protection through adversarial training. Raval \textit{et al.} \cite{raval2017protecting} propose a game-theoretic framework between a cloaking device and an attacker. Wu \textit{et al.}\cite{9207852} explicitly learn a degradation transformation for raw inputs. While such methods offer protection through software processing, the original, unprocessed video may already contain sensitive data vulnerable to theft.

\textbf{Hardware-based Methods.}
To address the risks of raw data leakage, physical-layer protection methods have emerged. Early works like Ryoo \textit{et al.}~\cite{Ryoo2017ExtremeLR} used low-resolution cameras, while Pittaluga \textit{et al.}~\cite{7298628} introduced defocused lenses. 
More recently, computational imaging techniques have advanced this field significantly. Wang \textit{et al.}~\cite{wang2019privacy} proposed coded aperture cameras that preserve motion features while hiding visual textures. Lensless imaging systems like DiffuserCam~\cite{antipa2017diffusercam} and FlatCam~\cite{asif2016flatcam} inherently produce measurements that appear unintelligible to humans, though such systems can often be partially inverted when the forward model is known or learned. To address this vulnerability, Cheng \textit{et al.} (DyPP)~\cite{cheng2024learning} proposed a dynamic privacy-preserving camera using time-varying PSFs, which offers theoretical guarantees against inversion attacks. Similarly, PrivHAR~\cite{PrivHAR} learns phase masks for action recognition. The importance of such hardware-level protection is further emphasized by recent benchmarks like EgoPrivacy~\cite{li2025egoprivacy}, which highlights the severe privacy risks in emerging domains like first-person vision.

While methods like DyPP~\cite{cheng2024learning} achieve rigorous protection through programmable optics, they often require specialized hardware control and expensive fabrication. In contrast, our Lens Privacy Sealing (LPS) explores a different direction: utilizing the \textit{stochastic scattering} of static, ultra-low-cost laminating films. Unlike engineered PSF methods that rely on deterministic optical coding, LPS introduces entropy-increasing degradation that is physically harder to invert. Our solution combines a novel device, dataset, and method (MSPNet) to balance privacy and accuracy in a scalable manner. We benchmark LPS against these representative methods under a unified protocol in Sec.~\ref{sec:tradeoff}.

\begin{figure*}[t]
    \centering
    \includegraphics[width=0.97\textwidth]{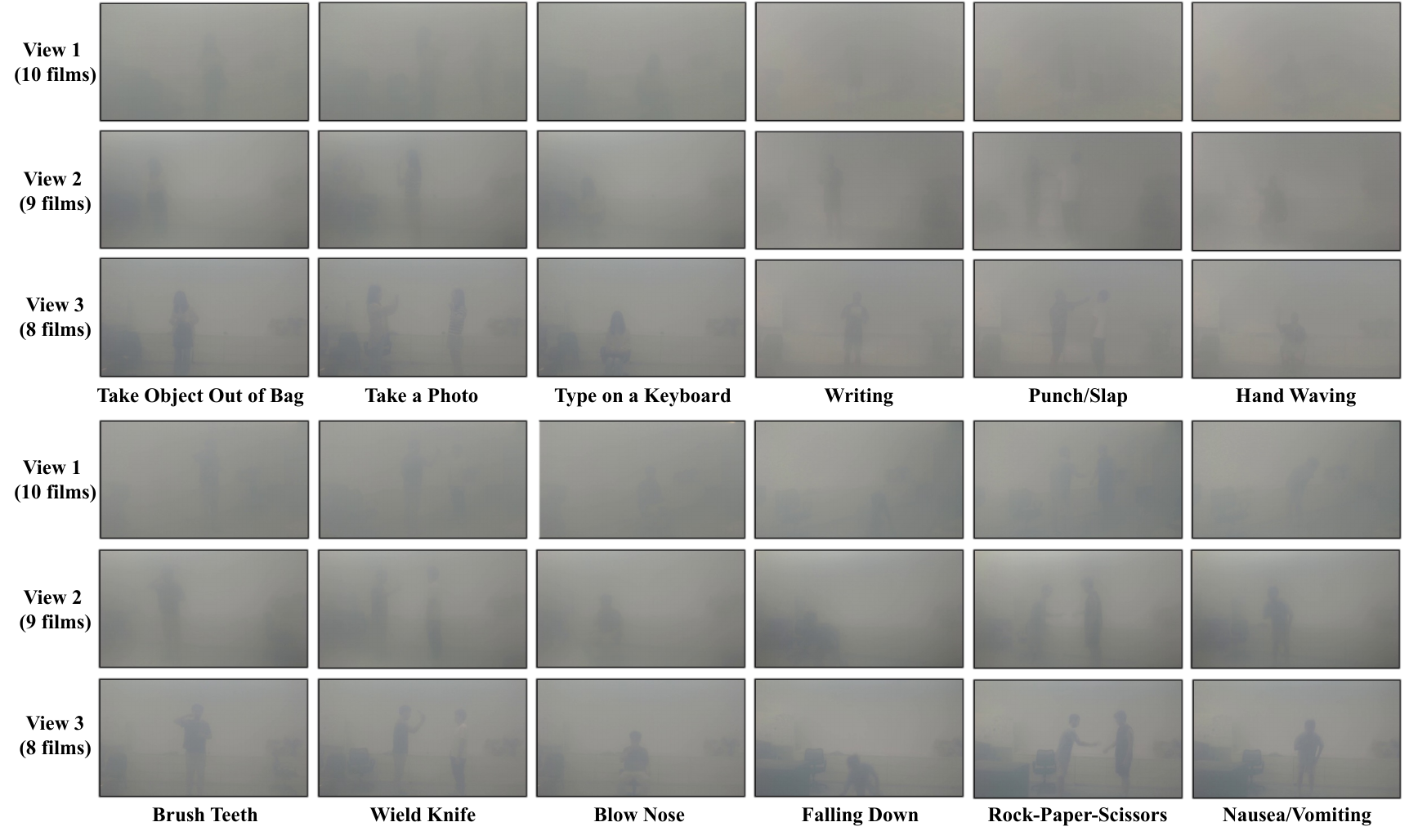}
    \vspace{-4mm}
    \caption{Samples of our P$^3$AR-PKU. The three camera views are captured using cameras covered with 8, 9, and 10 layers of laminating film, respectively.}
    \label{fig:pku60}
    \vspace{-3mm}
\end{figure*}

\section{Lens Privacy Sealing and The P$^3$AR Dataset}
\label{sec:dataset}

This section introduces our hardware solution and datasets: (1) \textbf{LPS}, a low-cost laminating film device for physical privacy protection; (2) \textbf{P$^3$AR-NTU}, a large-scale replay-and-capture dataset for systematic benchmarking; (3) \textbf{P$^3$AR-PKU}, a real-world dataset with multi-view and privacy attribute annotations. The combination of scale (NTU) and realism (PKU) enables comprehensive validation of privacy-utility trade-offs.

\subsection{Lens Privacy Sealing (LPS)}
Currently, hardware-level privacy protection solutions are limited and mainly rely on specialized cameras (e.g., defocused lenses~\cite{7298628}). However, these systems lack universality and are difficult to deploy in existing surveillance networks. To address this, we use an adjustable number of layers of 2R, 20C laminating film as a physical mask. 

\begin{figure}[t]
    \centering
    \includegraphics[width=0.48\textwidth]{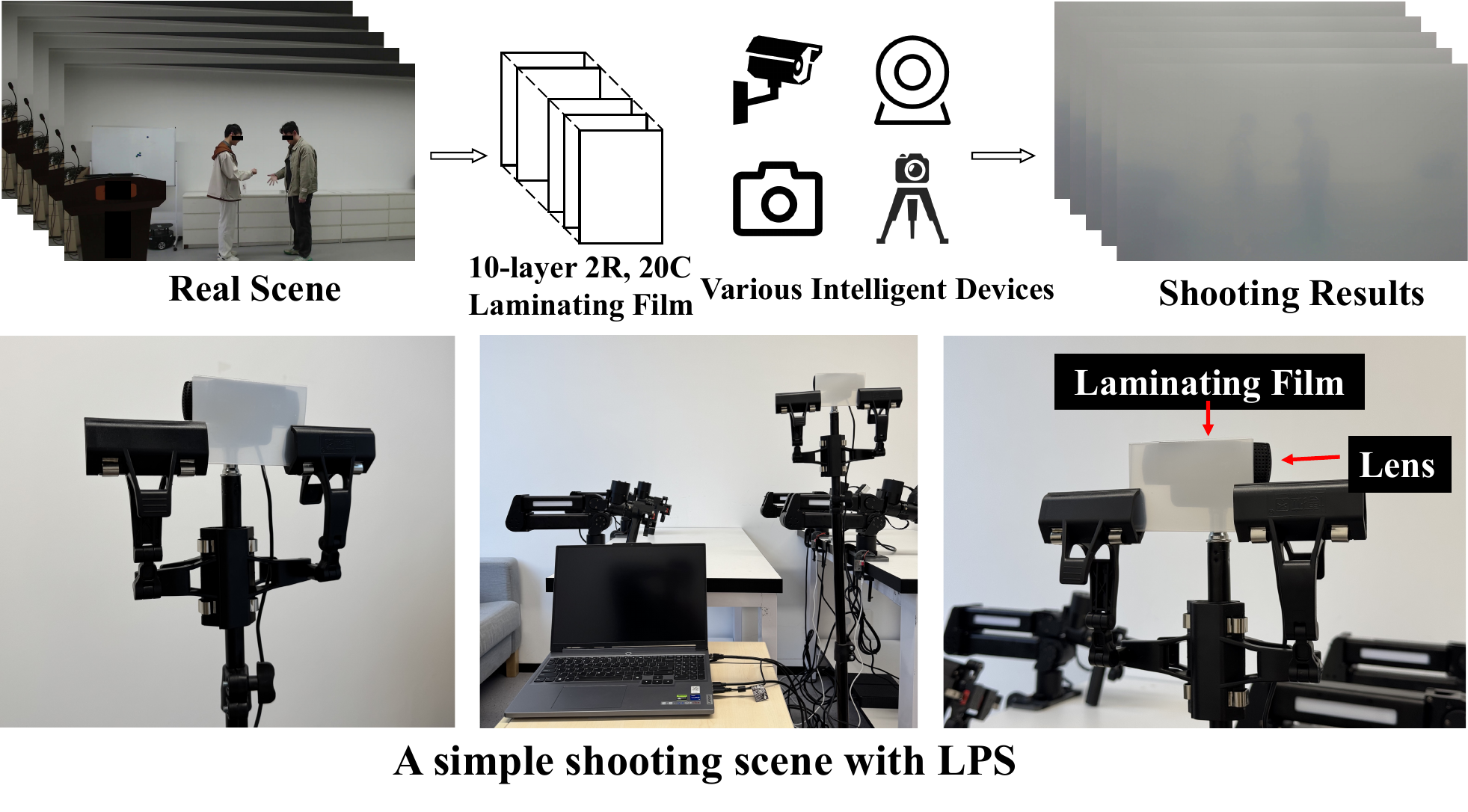}
    \vspace{-4mm}
    \caption{
    A workflow of our Lens Privacy Sealing. The adjustable laminating film can be customized by the user to cover the surface of various smart camera devices, achieving protection at the hardware level.}
    \label{fig:LPS}
    \vspace{-3mm}
\end{figure}

As shown in Fig.~\ref{fig:LPS}, LPS has several key characteristics: (1) \textbf{Low cost:} It is a relatively inexpensive material, making it suitable for large-scale use. (2) \textbf{Ease of use:} It can be quickly applied to objects of various shapes and sizes without complex equipment. (3) \textbf{Durability:} It is resistant to tearing, water, and dust, suitable for outdoor applications. (4) \textbf{Flexibility \& Adjustability:} The number of layers can be modified to achieve varying levels of obfuscation. 

An adversarial analysis of the obfuscated video captured using the LPS-equipped camera reveals that, after obfuscation, faces are not visible, skin tones are difficult to discern, and clothing colors are hard to identify.

\subsection{Dataset Construction}
To benchmark privacy-preserving action recognition, we constructed the P$^3$AR dataset using two complementary approaches: physically recapturing existing datasets and collecting real-world data.

\subsubsection{P$^3$AR-NTU: Physical Replay-and-Capture}
We constructed the large-scale P$^3$AR-NTU by physically recapturing the widely used NTU RGB+D 120 dataset.

\textbf{Rationale for Physical Capture:} We chose a physical replay-and-capture approach over software-based PSF simulation because real-world optical degradation involves complex non-uniform scattering and caustic effects. This methodology aligns with the rigorous hardware validation protocols established in recent state-of-the-art privacy research~\cite{cheng2024learning}, which emphasizes the necessity of constructing physical prototypes to validate privacy guarantees against inversion attacks, rather than relying solely on software simulation. Furthermore, evaluating on real-world data distributions is crucial for ensuring the validity of privacy assessment~\cite{li2025egoprivacy}.

The specific construction process involves three steps:
\begin{enumerate}
    \item \textbf{Physical Setup:} We utilized a controlled \textbf{monitor-replay protocol} in a \textbf{darkroom environment} to eliminate ambient light interference. The original NTU RGB+D videos were displayed on a \textbf{27-inch 4K monitor with a 144Hz refresh rate} to ensure high-fidelity playback.
    \item \textbf{Data Capture:} The displayed videos were recaptured by a \textbf{Logitech C920 PRO 1080p camera} equipped with the LPS device (specifically using \textbf{10 layers} of laminating film).
    \item \textbf{Post-Processing:} We performed \textbf{rigorous manual temporal alignment} to ensure precise synchronization between the degraded frames and the original labels.
\end{enumerate}

\subsubsection{P$^3$AR-PKU: Real-world Collection}
Since P$^3$AR-NTU is derived from displayed videos, it may not fully represent real-world lighting and depth conditions. To address this, we collected P$^3$AR-PKU using three cameras positioned at different angles in real-world settings. Each camera was equipped with 8, 9, or 10 layers of laminated film, respectively. This dataset includes 60 action categories (34 daily, 11 medical, 15 mutual actions) and annotates six privacy-sensitive attributes: personal ID, gender, hairstyle, eyeglasses, upper wear, and lower wear. Fig.~\ref{fig:pku60} visualizes samples under different viewpoints and film layers.

\subsection{Dataset Statistics and Comparison}
Table~\ref{tab:dataset1} compares the proposed P$^3$AR dataset with several other publicly available datasets. P$^3$AR-NTU consists of 114,480 video clips across 120 action categories performed by 106 individuals. It provides a large scale for training deep models. P$^3$AR-PKU, while smaller (4,500 clips), offers real-world diversity and privacy attribute annotations. The combination of these two subsets balances scale and realism.

\textbf{Temporal Specifications.} Each video clip in P$^3$AR has an average duration of 3--5 seconds, captured at 30 frames per second (FPS) with 1920$\times$1080 resolution. Following standard practice in human action recognition, we uniformly sample 8 frames per video and resize them to 224$\times$224 for model input. MSPNet achieves real-time inference with a latency of \textbf{35.72 ms per 8-frame clip} (batch size = 1, $\sim$28 clips/second) on a single NVIDIA RTX 4090 GPU, making it suitable for live surveillance applications.

\subsection{Image Quality Assessment}
We employ the following metrics to evaluate the degradation induced by LPS:
(1) \textbf{Structural Similarity Index Measure (SSIM)}~\cite{wang2004image}: evaluates image similarity by comparing brightness, contrast, and structural information.
(2) \textbf{Peak Signal-to-Noise Ratio (PSNR)}~\cite{hore2010image}: represents the signal-to-noise ratio.
(3) \textbf{Learned Perceptual Image Patch Similarity (LPIPS)}~\cite{zhang2018unreasonable}: measures perceptual similarity using deep features.
(4) \textbf{Mean Squared Error (MSE)}~\cite{gonzales1987digital}: quantifies the mean squared differences.

We conduct both qualitative and quantitative analyses on the NTU and P$^3$AR-NTU at both pixel and feature levels. The quantitative results in Table~\ref{tab:dataset2} demonstrate that LPS significantly degrades video image quality while altering the image distribution, thereby effectively enhancing privacy protection.

\begin{center}
\begin{figure*}[t]
\begin{center}
\includegraphics[width=1\textwidth]{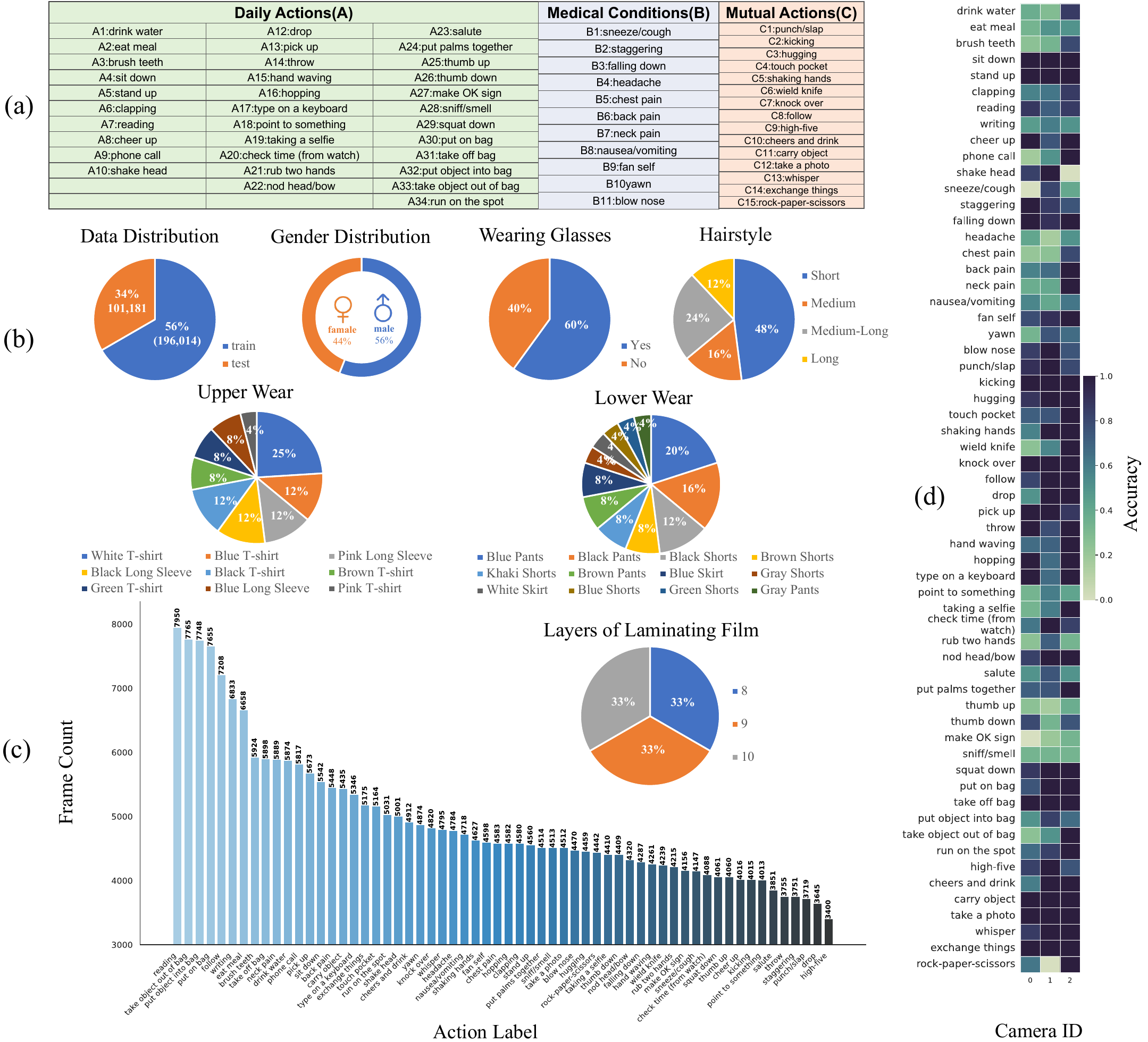}  
\end{center} 
\vspace{-2mm}
\caption{Details of the P$^3$AR-PKU. (a) Action Labels: Includes three categories: Daily Actions, Medical Conditions, and Mutual Actions. (b) Privacy Attributes: Data distribution based on attributes such as gender, hairstyle and clothing of the participants, as well as the train/test split ratio. (c) Frame Statistics: The number of video frames corresponding to each action label in the P$^3$AR-PKU. (d) Recognition Accuracy: The accuracy of action recognition for each action across three different viewpoints in the P$^3$AR-PKU, evaluated using the MSPNet method.
}
\label{fig:pku60_app}
\vspace{-4mm}
\end{figure*}
\end{center}

\renewcommand{\arraystretch}{1}
\renewcommand{\arraystretch}{1.3}
\begin{table}[t]
\centering
\caption{Quantitative analysis of video quality degradation in P$^3$AR-NTU at original and resized (224$\times$224) resolutions.}
\vspace{-2mm}
\begin{tabular}{ccccc}
\toprule
\textbf{Dataset Version} & \textbf{SSIM $\downarrow$}& \textbf{PSNR $\downarrow$} & \textbf{LPIPS $\uparrow$} & \textbf{MSE $\uparrow$} \\ 
\hline
P$^3$AR-NTU         & 0.634 & 27.95  &  0.454  & 101.76     \\ 
\hline
P$^3$AR-NTU (224, 224)          & 0.447 & 27.94  &0.711  &102.22       \\ 
\bottomrule
\end{tabular}
\vspace{-1em}
\label{tab:dataset2}
\end{table}
\renewcommand{\arraystretch}{1}

\section{Method}
\label{sec:method}

This section presents MSPNet in three parts: the imaging model $\mathbf{y} = \mathcal{H}(\mathbf{x})+\mathbf{n}$ and key assumptions (Sec.~\ref{sec:formulation}), IFNS for suppressing static scattering via frame differencing, and CFSA with CLIP prompting for multi-scale temporal modeling and semantic alignment.

\subsection{Problem Formulation}
\label{sec:formulation}
To clarify the link between the optical degradation induced by LPS and our network design, we first formulate the imaging process. Let $\mathbf{x}_t \in \mathbb{R}^{H \times W \times 3}$ denote the latent clean scene at time $t$, and $\mathbf{y}_t$ denote the captured degraded frame. While actual multi-layer scattering involves complex nonlinear effects, we adopt a first-order linear approximation for tractable analysis:
\begin{equation}
    \mathbf{y}_t = \mathcal{H}(\mathbf{x}_t) + \mathbf{n}_t,
\end{equation}
where $\mathcal{H}$ represents the scattering operator (approximated as a convolution with the film's Point Spread Function), and $\mathbf{n}_t$ represents sensor noise. We assume the scattering medium is time-invariant and the background scene $\mathbf{x}^{bg}$ is static ($\mathbf{x}_t^{bg} \approx \mathbf{x}_{t-1}^{bg}$).

Based on the linearity of $\mathcal{H}$, the frame difference used in our Inter-Frame Noise Suppressor (IFNS) can be expressed as:
\begin{equation}
    \Delta \mathbf{y} = \mathbf{y}_t - \mathbf{y}_{t-1} = \mathcal{H}(\mathbf{x}_t - \mathbf{x}_{t-1}) + (\mathbf{n}_t - \mathbf{n}_{t-1}).
\end{equation}
Since the static background components cancel out ($\mathbf{x}_t - \mathbf{x}_{t-1} \approx \mathbf{x}_t^{motion}$), this operation effectively removes the static scattering haze and background clutter. This theoretical insight justifies our design of IFNS, which allows the network to focus purely on the motion dynamics modulated by the scattering kernel, rather than struggling with global degradation.

\textbf{Assumptions and Limitations.}
Our formulation relies on two key assumptions: (1) the scattering operator $\mathcal{H}$ is approximately time-invariant within a short action clip, and (2) the background scene is mostly static ($\mathbf{x}_t^{bg} \approx \mathbf{x}_{t-1}^{bg}$). These assumptions hold well in typical fixed-camera surveillance and smart-home deployments. We systematically evaluate robustness under assumption violations---including dynamic backgrounds, varying illumination, and camera motion---in Section~\ref{sec:robustness}, demonstrating that MSPNet maintains strong performance under most challenging conditions, with camera motion being the primary limitation.

\textbf{First-order Approximation vs.\ Actual Complexity.}
It is important to distinguish the requirements for IFNS from those for image reconstruction. IFNS only requires that \textit{scattering patterns remain temporally stable within a short clip and backgrounds are approximately static}---it does not require $\mathcal{H}$ to be strictly invertible or spatially uniform. In contrast, deconvolution-based reconstruction demands that the inverse problem be well-posed. As shown in Section~\ref{sec:attack}, reconstruction attacks fail not because the first-order linear model is invalid, but because (i) high-frequency information is severely attenuated below the noise floor, (ii) noise amplification makes the inverse ill-conditioned, and (iii) spatial variations and nonlinear effects further degrade inversion accuracy. Thus, the same optical system that supports effective frame differencing for action recognition simultaneously resists inversion attacks due to the ill-posed nature of the reconstruction problem.

\subsection{Overview of MSPNet}
\textbf{Motivation for Single-stage Design.}
Traditional HAR pipelines follow a two-stage paradigm: detecting humans first, then analyzing actions. However, under severe optical scattering induced by LPS, the intermediate bounding-box representation becomes unreliable due to detection jitter and cropping artifacts. In contrast, single-stage methods that directly process full video frames can bypass these unstable intermediate representations. To address the specific challenges of LPS---heavy scattering noise and relatively small subject regions---we introduce two specialized modules: the Inter-Frame Noise Suppressor (IFNS) and the Cross-Frame Semantic Aggregator (CFSA).

\begin{figure*}[t]
    \centering
    \includegraphics[width=1\textwidth]{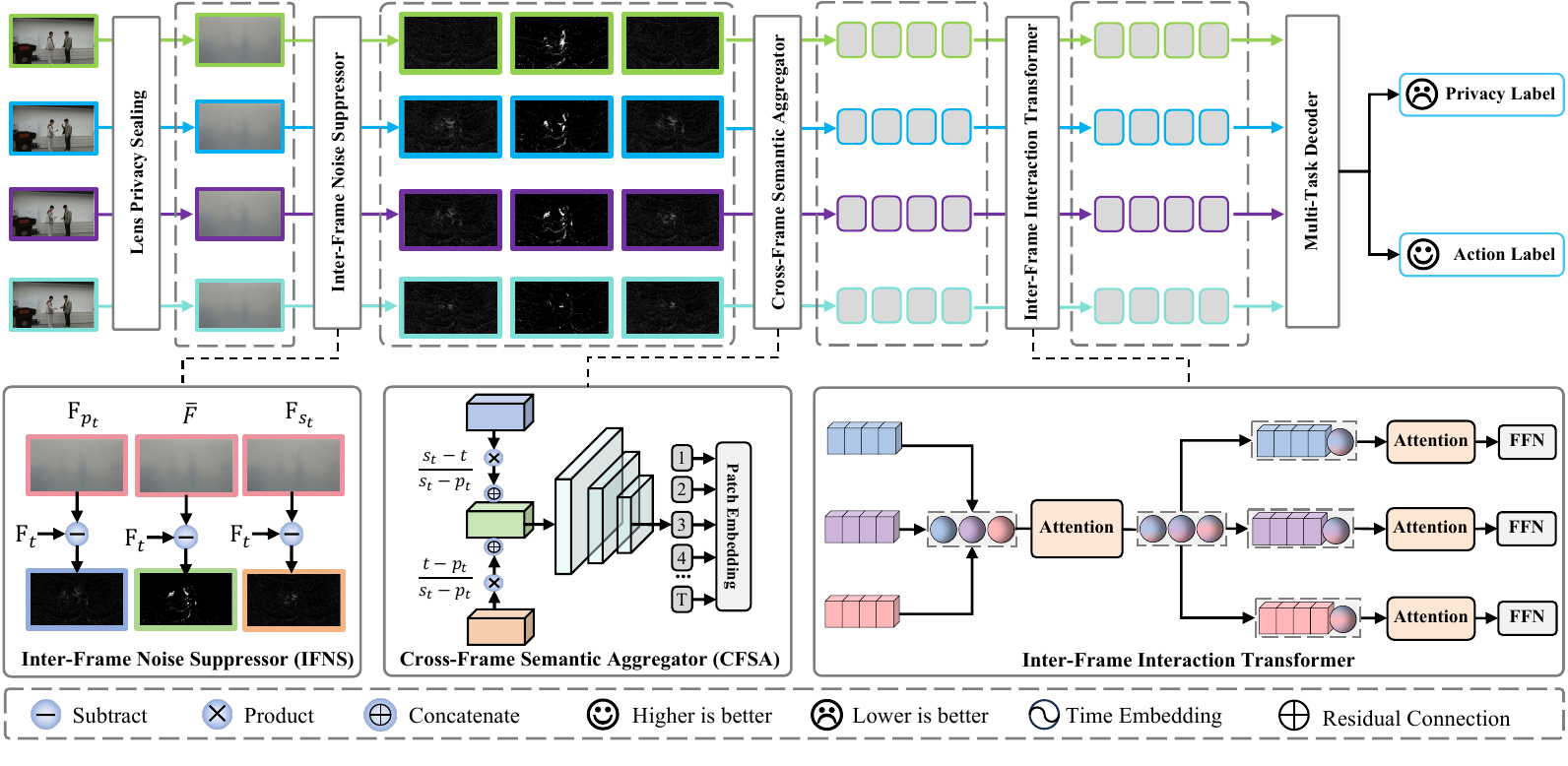}
    \vspace{-4mm}
    \caption{Pipeline of MSPNet. The details of IFNS, CFSA, and the Inter-Frame Interaction Transformer. Specifically, IFNS and CFSA are illustrated with respect to the $t$-th frame. $F$ represents the video frame set in a sample, and $\bar{F}$ denotes the average frame of the sample. The indices $p_t$, $t$, and $s_t$ in the figure refer to the previous frame, the current frame, and the subsequent frame, respectively.} 
    \label{fig:pipeline}
    \vspace{-4mm}
\end{figure*}

\textbf{Architecture Overview.}
As shown in Fig.~\ref{fig:pipeline}, MSPNet comprises three stages. First, the Inter-Frame Noise Suppressor (IFNS) computes residuals between the current frame and past frames, future frames, and the average frame. Second, the Cross-Frame Semantic Aggregator (CFSA) fuses these residuals based on temporal distances. Third, Privacy Motion Recognition uses CFSA output to generate video-specific text prompts, and the recognition result is obtained via cosine similarity. The input consists of eight uniformly sampled frames (224$\times$224), along with preceding, subsequent, and average frames for residual computation. IFNS and CFSA perform residual processing and fusion. The Inter-frame Interaction Transformer (IIT) handles single-frame processing and cross-frame exchange. Finally, the Spatio-temporal Fusion Transformer converts frame-level representations into video-level representations.
For the pre-trained text encoder, text descriptions of all action categories in the dataset are provided as input. Within the text encoder, all action labels are combined with the video-level representations output by the Spatio-temporal Fusion Transformer. This combination leverages video content information to enhance the generation of text prompts, producing video-specific prompts aligned with the content. Finally, cosine similarity is calculated between each video and all candidate action categories (text prompts), allowing the model to determine the most likely action category for the video by selecting the category with the highest similarity as the recognition result. 

\subsection{Inter-Frame Noise Suppressor}
We observed that the scattering pattern induced by laminating film remains highly consistent across frames within the same video. This temporal consistency motivates our Inter-Frame Noise Suppressor (IFNS), which leverages inter-frame grayscale residuals to suppress static scattering while preserving motion contours. By computing frame differences, IFNS effectively removes the background degradation and minimizes data redundancy, retaining the critical motion details essential for HAR.

Previous methods utilizing inter-frame residuals primarily focused on motion information between adjacent frames, but this information is often insufficient, as some behavioral patterns only emerge over longer time scales. To address this, IFNS additionally computes the residual with respect to the average frame. An added benefit of using the average frame is that the static background components are retained during the averaging process, while the moving foreground components are smoothed out or even canceled due to their positional and morphological changes. This entire process requires no complex modeling, making it particularly suitable for degraded videos. Moreover, the visualized results of IFNS clearly show that while emphasizing the contours of the moving subject, the information about the subject's gender, clothing, skin tone, and body weight becomes difficult to distinguish.

Consider a video sequence \( V = \{F_t\}_{t=1}^T \), where \( F_t \) represents a single frame at time \( t \). Each \( F_t \) is a two-dimensional grayscale image with pixel values \( F_t(x, y) \) defined on the image plane, where \( (x, y) \) denotes the pixel position. The average frame \( \bar{F} \) of the video is 
$\bar{F}(x, y) = \frac{1}{T} \sum_{t=1}^{T} F_t(x, y)$.
For each current frame \( F_{t} \), we first compute the frame difference with the previous frame \( F_{p_t} \), the frame difference with the subsequent frame \( F_{s_t} \), and the frame difference with the average frame \( \bar{F} \). Here, \( p_t \) and \( s_t \) represent the selected frames before and after the present frame, respectively, and they are not necessarily adjacent frames. The specific frame difference operations are defined as:

\begin{equation}
\begin{aligned}
D_t^{\text{p}}(x, y) &= \lvert F_{t}(x, y) - F_{p_t}(x, y) \rvert, \\
D_t^{\text{s}}(x, y) &= \lvert F_{t}(x, y) - F_{s_t}(x, y) \rvert, \\
D_t^{\text{m}}(x, y) &= \lvert F_{t}(x, y) - \bar{F}(x, y) \rvert.
\end{aligned}
\end{equation}

In this formula, \( p_t \) and \( s_t \) can be chosen with different frame intervals based on specific needs. This flexibility allows us to select appropriate frame intervals when calculating the frame difference based on the temporal characteristics or motion patterns in the video, thereby improving the model's ability to recognize behaviors over different time scales. Notably, to ensure the consistency of the channel count after fusion, all three differential images are single-channel grayscale images.

\subsection{Cross-Frame Semantic Aggregator}
After obtaining the differential images through IFNS, we propose the Cross-Frame Semantic Aggregator (CFSA) to better leverage these images as valuable temporal cues, effectively aggregating cross-frame semantic information. CFSA performs weighted operations on the preceding and subsequent frame differences based on the interval between frames. The final feature \( F_t^{\text{CFSA}}(x, y) \) is obtained by concatenating these weighted frame differences with the mean frame difference, as expressed by the following equation:
\begin{equation}
\text{Concat}\left( \frac{s_t - t}{s_t - p_t} D_t^{\text{p}}(x, y), \frac{t - p_t}{s_t - p_t} D_t^{\text{s}}(x, y), D_t^{\text{m}}(x, y) \right).
\end{equation}

Finally, \( F_t^{\text{CFSA}}(x, y) \) is processed through a 3 × 3 convolutional module to obtain the final representation \( \Psi_t(x, y) \).

\subsection{CLIP-based Model Architecture}
To exploit rich semantic information for action recognition, we extend contrastive language-image pretraining to video-level recognition. Our model consists of a video encoder $E_v$ and a text encoder $E_t$. Given the feature $\mathbf{F}_v$ from CFSA and label $y$, the video encoder produces $\mathbf{v} = E_v(\mathbf{F}_v)$, while the text encoder generates $\mathbf{t} = E_t(y)$. Video-specific Prompting then enhances $\mathbf{t}$ using $\mathbf{v}$ to produce $\mathbf{t}_v$. The model is trained to maximize similarity between matched video-text pairs.

\textbf{Video Encoder.}
The video encoder comprises Patch Embedding, Inter-frame Interaction Transformer (IIT), and Spatio-temporal Fusion Transformer. As illustrated in Fig.~\ref{fig:pipeline}, IIT consists of Temporal Fusion Attention and Self-Propagation Attention. Each frame's features are extracted into message tokens through a linear layer, followed by multi-head self-attention for cross-frame information exchange. Temporal Fusion Attention captures global spatio-temporal dependencies, while Self-Propagation Attention combines intra-frame features with cross-frame features for a full spatio-temporal receptive field.

The Spatio-temporal Fusion Transformer integrates frame representations with temporal position encoding to generate the final video representation:
\begin{equation}
\mathbf{v} = \text{Pooling}(\text{Transformer}(\mathbf{H} + \mathbf{P}_t))
\end{equation}
where $\mathbf{H}$ represents all frame representations and $\mathbf{P}_t$ is the temporal position encoding.

\textbf{Text Encoder and Video-specific Prompting.}
The text encoder transforms label $y$ into representation $\mathbf{t}$. To align text with video content, we design a video-specific prompt generator that dynamically adapts the text representation:
\begin{equation}
\mathbf{t}_v = \mathbf{t} + \alpha \cdot \text{FFN}(\text{MHSA}(\mathbf{t}, \frac{1}{T} \sum_{i=1}^{T} \mathbf{h}_i)),
\end{equation}
where $T$ is the number of frames, $\mathbf{h}_i$ is the $i$-th frame representation, and $\alpha$ is the fusion coefficient.

\textbf{Training Objective.}
We adopt contrastive learning to align video and text representations:
\begin{equation}
\mathcal{L} = -\log \frac{\exp(\text{sim}(\mathbf{v}, \mathbf{t}_v) / \tau)}{\sum_{k=1}^{N} \exp(\text{sim}(\mathbf{v}, \mathbf{t}_k) / \tau)},
\end{equation}
where $\tau$ is the temperature parameter, $N$ is the batch size, and $\text{sim}(\cdot, \cdot)$ denotes cosine similarity.

\section{Experiment}
\label{sec:experiment}

This section is organized as follows: we first describe evaluation metrics and implementation details (Sec.~\ref{sec:metrics}--\ref{sec:impl}), then compare with related methods and analyze privacy-utility trade-offs (Sec.~\ref{sec:comparison}--\ref{sec:tradeoff}), followed by robustness validation against reconstruction attacks and environmental variations (Sec.~\ref{sec:attack}--\ref{sec:robustness}), and finally present ablation studies (Sec.~\ref{sec:ablation}).

\subsection{Evaluation Metrics}
\label{sec:metrics}
We use two primary metrics to evaluate our system: (1) \textbf{Subject ID Accuracy ($ACC_S$)}: the identification rate of the subject's identity. A lower $ACC_S$ indicates better privacy protection. To ensure reliable privacy risk assessment, well-calibrated classifiers~\cite{Cheng_2022_CVPR} are important for accurate probability estimation. (2) \textbf{Action Recognition Accuracy ($ACC_{act}$)}: the accuracy of human action recognition, which reflects the utility of the HAR system. The goal is to maintain high action recognition accuracy while achieving low $ACC_S$. The privacy attributes in the P$^3$AR-PKU are also evaluated using accuracy.

\renewcommand{\arraystretch}{1.3}
\begin{table}[t]
\centering
\caption{
\textbf{Comparison of Top-1 Action Recognition Accuracy (\%) on P$^3$AR-NTU.} 
``\textbf{Double}'' denotes the two-stage pipeline (Human Detection $\to$ Action Recognition), which is sensitive to detection errors and cropping artifacts under severe LPS degradation. 
``\textbf{Single}'' denotes the baseline single-stage pipeline (Direct Action Recognition). 
``\textbf{with IFNS \& CFSA}'' indicates our proposed MSPNet. 
\textbf{Bold} indicates the best performance.
}
\footnotesize
\resizebox{\columnwidth}{!}{
\begin{tabular}{r|c|c|c}
\toprule
\textbf{Method} & \textbf{Double} & \textbf{Single} & \textbf{with IFNS \& CFSA} \\ 
\hline
TSN-ResNet50 (ECCV 2016)~\cite{tsn} & 19.17 & 20.21 & 34.40 \\
TSN-DenseNet161 (ECCV 2016)~\cite{tsn} & 20.83 & 21.43 & 31.66 \\
SlowFast-ResNet50 (ICCV 2019)~\cite{slowfast} & 20.88 & 22.16 & 59.27 \\
SlowOnly-ResNet50 (ICCV 2019)~\cite{slowfast} & 25.90 & 24.88 & 55.75 \\
TSM-ResNet50 (ICCV 2019)~\cite{tsm} & 23.18 & 26.12 & 59.68 \\
TimeSformer-space (ICML 2021)~\cite{timesformer} & 24.40 & 21.05 & 60.22 \\
MViT V2-S (CVPR 2022)~\cite{mvitv2} & 29.98 & 31.51 & 63.24 \\
VideoSwin-Swin-B (CVPR 2022)~\cite{videoswin} & 34.70 & 33.79 & 64.57 \\
VideoMAE-ViT-B (NeurIPS 2022)~\cite{videomae} & 28.62 & 32.07 & 62.05 \\
VideoMAE V2-ViT-B (CVPR 2023)~\cite{videomaev2} & 31.09 & 35.21 & 63.65 \\
UniFormer V2-B/16 (ICCV 2023)~\cite{uniformerv2} & 31.85 & 34.26 & 64.92 \\
InternVideo2-B/14 (ECCV 2024)~\cite{wang2025internvideo2} & 36.58 & 36.93 & 65.97 \\
\midrule
\textbf{MSPNet (Ours)} & \textbf{37.35} & \textbf{38.38} & \textbf{67.90} \\
\bottomrule
\end{tabular}}
\label{tab:Comparison}
\vspace{-4mm}
\end{table}
\renewcommand{\arraystretch}{1}

\subsection{Implementation Details}
\label{sec:impl}
The capture system consists of a Logitech C920 PRO 1080p camera covered with an adjustable number of layers of 2R, 20C laminating film. Additionally, the number of laminating film layers can be adjusted based on specific application requirements, rather than being limited to ten layers. If enhanced privacy protection is needed, additional layers can be applied. Conversely, if higher human action recognition accuracy is desired, the number of layers can be reduced. 

Our method is implemented using the PyTorch 1.8 framework with CUDA 11.2 and trained on two NVIDIA 4090 GPUs. We use Stochastic Gradient Descent (SGD) optimizer with a mini-batch size of $64$, a momentum of $0.9$, and a weight decay of $5e^{-4}$ to train the model for 200 epochs. The initial learning rate was set to $0.05$ with a decay of $0.1$. In the P$^3$AR-PKU assessment, the training group and the test group are partitioned based on Subject, with a ratio of 2:1.

\subsection{Comparison with Related Methods}
\label{sec:comparison}
\subsubsection{Quantitative Comparison on P$^3$AR-NTU}
Table~\ref{tab:Comparison} compares MSPNet with 12 state-of-the-art methods on P$^3$AR-NTU. We evaluate three settings: (1) \textbf{Double} (two-stage: detection then recognition), (2) \textbf{Single} (direct recognition without detection), and (3) \textbf{with IFNS \& CFSA} (our full pipeline).

Due to the severe optical degradation from LPS, two-stage methods suffer from unreliable detection, leading to cascading errors. Single-stage methods perform better by avoiding detection failures, but still struggle without specialized modules for degraded video. Our IFNS and CFSA modules address this: IFNS suppresses scattering noise through frame differencing while preserving motion contours, and CFSA aggregates cross-frame semantics with dynamic temporal weighting. As shown in Table~\ref{tab:Comparison}, integrating IFNS and CFSA nearly doubles the accuracy for all methods, with MSPNet achieving the highest accuracy of 67.90\%. Notably, these modules can be integrated into existing HAR architectures.

\renewcommand{\arraystretch}{1.3} 
\begin{table*}[htb]
\centering
\caption{
\textbf{Quantitative Comparison on Real-world P$^3$AR-PKU Dataset.} 
We evaluate both Utility (Action Recognition) and Privacy (Attribute Leakage). 
$\uparrow$ indicates higher is better (for Utility), while $\downarrow$ indicates lower is better (for Privacy). 
\textbf{Bold} indicates the best performance (Highest Action Acc or Lowest Attribute Acc).
}
\vspace{-2mm}
\resizebox{\textwidth}{!}{
\begin{tabular}{r|c|cccccc}
\toprule
\multirow{1}{*}{\textbf{}} & \textbf{Action category} & \multicolumn{6}{c}{\textbf{Privacy attributes (Lower is Better)}} \\ 
\cline{2-8}
\textbf{Method} & \textbf{$ACC_{act}\uparrow$} & \textbf{$ACC_S\downarrow$} & \textbf{$Acc_{gen}\downarrow$} & \textbf{$Acc_{hair}\downarrow$} & \textbf{$Acc_{gls}\downarrow$} & \textbf{$Acc_{upper}\downarrow$} & \textbf{$Acc_{lower}\downarrow$} \\
\hline
TSN-ResNet50 (ECCV 2016)~\cite{tsn} & 43.84 & 8.53 & 71.79 & 59.44 & 67.85 & 28.38 & 11.90 \\
TSN-DenseNet161 (ECCV 2016)~\cite{tsn} & 39.54 & 6.15 & 68.61 & 60.20 & 66.67 & 25.33 & 10.78 \\
TRN-ResNet50 (ECCV 2018)~\cite{trn} & 51.85 & 10.52 & 65.78 & 60.42 & 68.42 & 20.47 & 13.47  \\
SlowFast-ResNet50 (ICCV 2019)~\cite{slowfast} & 63.84 & \textbf{5.25} & 65.97 & 54.19 & 66.56 & 21.73 & 9.29 \\
SlowOnly-ResNet50 (ICCV 2019)~\cite{slowfast} & 60.40 & 5.36 & 66.84 & 53.21 & 68.05 & 22.47 & \textbf{7.68}  \\
TSM-ResNet50 (ICCV 2019)~\cite{tsm} & 63.33 & 8.82 & 69.33 & 58.14 & 70.33 & 24.58 & 10.32  \\
TimeSformer-spaceOnly (ICML 2021)~\cite{timesformer} & 64.33 & 10.67 & 67.27 & 60.05 & 70.14 & 20.91 & 10.91 \\
MViT V2-S (NeurIPS 2022)~\cite{mvitv2} & 67.69 & 8.73 & 66.67 & 60.40 & 72.55 & 18.21 & 10.67 \\
VideoSwin-Swin-B (CVPR 2022)~\cite{videoswin} & 71.73 & 6.31 & 72.68 & 55.79 & \textbf{59.47} & 23.68 & 7.93 \\
VideoMAE-ViT-B (NeurIPS 2022)~\cite{videomae} & 69.45 & 9.42 & 69.67 & 55.83 & 67.78 & 23.33 & 12.85 \\
VideoMAE V2-ViT-B (CVPR 2023)~\cite{videomaev2} & 71.52 & 9.26 & 73.16 & \textbf{51.80} & 69.84 & 17.95 & 12.67 \\
UniFormer V2-B/16 (ICCV 2023)~\cite{uniformerv2} & 72.29 & 5.83 & 72.58 & 54.41 & 70.25 & \textbf{17.33} & 10.72 \\
InternVideo2-B/14 (ECCV 2024)~\cite{wang2025internvideo2} & 73.25 & 5.67 & \textbf{65.33} & 57.13 & 68.65 & 25.31 & 11.87 \\
\hline
\textbf{MSPNet (Ours)} & \textbf{74.83} & 5.67 & 67.50 & 54.83 & 69.91 & 24.91 & 9.76 \\
\bottomrule
\end{tabular}
}
\vspace{-4mm}
\renewcommand{\arraystretch}{1}
\label{tab:real}
\end{table*}

Thanks to the model's capability to extract spatio-temporal information, our proposed MSPNet achieves an accuracy of 67.90\%, the highest among all methods. Compared to InternVideo2~\cite{wang2025internvideo2}, MSPNet shows a 1.93\% improvement, demonstrating the effectiveness of IFNS \& CFSA for privacy-preserving action recognition.

\subsubsection{Quantitative Comparison on P$^3$AR-PKU}
We also evaluate on the real-world P$^3$AR-PKU dataset. Table~\ref{tab:real} presents the performance of MSPNet and other methods. Since P$^3$AR-PKU involves multiple privacy attributes, we follow multi-label evaluation practices~\cite{Cheng_2024_CVPR} for comprehensive privacy assessment. The low recognition accuracy of all privacy attributes verifies LPS effectiveness in privacy protection, while the excellent action recognition accuracy (74.83\%) demonstrates the superiority of our recognition framework. Fig.~\ref{fig:pku60_app}(d) illustrates the per-action accuracy across three viewpoints.

\subsection{Privacy-Utility Trade-off Analysis}
\label{sec:tradeoff}

\textbf{Defining Performance Metrics.} We define two key concepts for evaluating privacy-preserving action recognition: (1) \textit{Ideal Level} refers to the action recognition accuracy achievable on clean, unprotected data, serving as the theoretical upper bound; (2) \textit{Acceptable Trade-off} is achieved when a method lies on the Pareto frontier, meaning no other method achieves both higher utility and lower privacy risk. For NTU RGB+D 120 with 106 subjects, random-guess $ACC_S$ is $\frac{1}{106} \approx 0.94\%$; we consider $ACC_S < 15\%$ as strong privacy protection and $ACC_{act} > 65\%$ as practical utility.

To rigorously evaluate LPS and justify its position as a benchmark for physical privacy-preserving action recognition, we compare it with representative software and hardware-based privacy methods on NTU RGB+D 120.

\subsubsection{Compared Methods and Protocol}
\textbf{Software Baselines:} (1) Gaussian Blur ($\sigma$=10): simulates traditional optical defocus; (2) Low Resolution ($\times$8): applies downsampling to simulate low-resolution capture.

\textbf{Hardware Methods:} (1) Coded Aperture~\cite{wang2019privacy}: uses a pseudo-random mask with 50\% transmittance; (2) DiffuserCam~\cite{antipa2017diffusercam}: simulates lensless imaging with a random speckle PSF computed via Angular Spectrum Propagation; (3) DyPP~\cite{cheng2024learning}: the state-of-the-art dynamic privacy-preserving camera using time-varying PSFs parametrized by 350 Zernike coefficients.

All baselines are applied to the RGB videos of NTU RGB+D 120 to generate privacy-protected inputs. For software methods (Blur, Low Resolution), we apply the degradation directly to clean NTU frames. For hardware methods (Coded Aperture, DiffuserCam, DyPP), we simulate optical degradation using convolution with the corresponding PSFs. For LPS, we use the physically captured P$^3$AR-NTU, which is a replay-and-capture version of NTU RGB+D 120 with identical action labels.

We evaluate along two dimensions: \textbf{Utility} (Action Recognition Accuracy using MSPNet, Cross-Subject split) and \textbf{Privacy Risk} (Subject ID Accuracy $ACC_S$ using UniFormer V2, Cross-View split). To ensure fairness, all methods use the same backbone, data augmentation, and frame sampling strategy. Since most methods lack open-source implementations, we faithfully reproduced them based on original papers and released all code in our GitHub repository.

\subsubsection{Quantitative Results}
\begin{figure}[t]
    \centering
    \begin{tikzpicture}
        \begin{axis}[
            width=0.95\columnwidth,
            height=6cm,
            grid=major,
            x dir=reverse, 
            xlabel={Subject ID Accuracy ($ACC_S$, \%) $\downarrow$ (Privacy Risk)},
            ylabel={Action Recognition Accuracy (\%) $\uparrow$},
            xmin=0, xmax=100,
            ymin=0, ymax=100,
            font=\small,
            tick label style={font=\small},
            label style={font=\small},
            legend style={at={(0.02,0.02)},anchor=south west, font=\footnotesize},
        ]
        
        \addplot[
            color=red,
            mark=square*,
            thick,
            nodes near coords, 
            point meta=explicit symbolic, 
            every node near coord/.append style={anchor=south west, font=\scriptsize, color=red, xshift=2pt}
        ]
        coordinates {
            (98.36, 79.52) [0 Layers]
            (40.05, 75.29) [6 Layers]
            (19.57, 71.03) [8 Layers]
            (9.05, 67.90) [10 Layers]
        };
        \addlegendentry{Ours (LPS + MSPNet)}

        \addplot[
            color=blue,
            mark=triangle*,
            only marks,
            mark size=2.5pt,
            nodes near coords,
            point meta=explicit symbolic,
            every node near coord/.append style={anchor=south, font=\scriptsize, color=blue, yshift=1pt}
        ]
        coordinates {
            (87.91, 73.41) [Gaussian]
        };
        \addlegendentry{Software Baselines}
        
        \addplot[
            color=blue,
            mark=triangle*,
            only marks,
            mark size=2.5pt,
            nodes near coords,
            point meta=explicit symbolic,
            every node near coord/.append style={anchor=north, font=\scriptsize, color=blue, yshift=-1pt},
            forget plot
        ]
        coordinates {
            (83.16, 72.28) [LowRes]
        };
        
        \addplot[
            color=green!60!black,
            mark=diamond*,
            only marks,
            mark size=2.5pt,
            nodes near coords,
            point meta=explicit symbolic,
            every node near coord/.append style={anchor=south, font=\scriptsize, color=green!60!black, yshift=3pt}
        ]
        coordinates {
            (3.94, 27.30) [Coded Aperture]
            (4.26, 16.28) [Diffuser]
            (19.83, 39.52) [DyPP]
        };
        \addlegendentry{Hardware SOTA}
            
        \end{axis}
    \end{tikzpicture}
    \vspace{-7mm}
    \caption{\textbf{Privacy-Utility Trade-off Comparison.} The LPS curve (red, 0--10 layers) forms a Pareto Frontier superior to software baselines (blue) and hardware methods (green). The x-axis is reversed: lower $ACC_S$ (better privacy) appears on the right. An ideal method maximizes $ACC_{act}$ while minimizing $ACC_S$, corresponding to the \textbf{top-right} region. LPS achieves both strong privacy ($ACC_S$=9.05\%) and high utility ($ACC_{act}$=67.90\%).}
    \label{fig:tradeoff}
    \vspace{-4mm}
\end{figure}
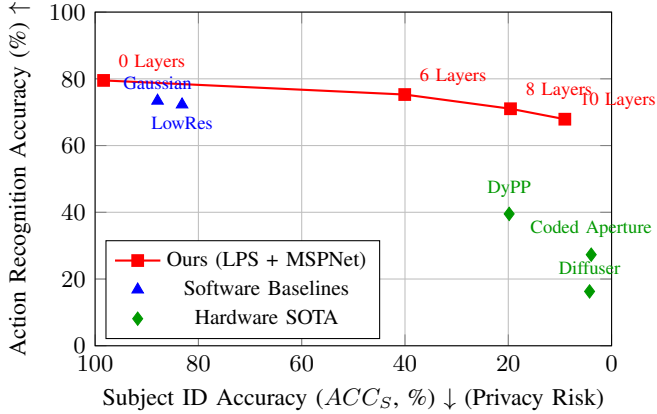

Table~\ref{tab:privacy_comparison} and Fig.~\ref{fig:tradeoff} present the comparison results.

\begin{table}[t]
\centering
\caption{\textbf{Privacy-Utility Comparison with SOTA Methods on NTU RGB+D 120.} LPS achieves the best trade-off: high utility with low privacy risk. $\uparrow$: higher is better; $\downarrow$: lower is better.}
\label{tab:privacy_comparison}
\renewcommand{\arraystretch}{1.2}
\resizebox{\columnwidth}{!}{
\begin{tabular}{l|c|c|c}
\toprule
\textbf{Method} & \textbf{Type} & \textbf{Utility ($ACC_{act} \uparrow$)} & \textbf{Privacy ($ACC_S \downarrow$)} \\ 
\midrule
Clean RGB (Baseline) & - & 79.52\% & 98.36\% \\ 
\midrule
Gaussian Blur ($\sigma$=10) & Software & 73.41\% & 87.91\% \\
Low Resolution ($\times$8) & Software & 72.28\% & 83.16\% \\ 
\midrule
Coded Aperture~\cite{wang2019privacy} & Hardware & 27.30\% & 3.94\% \\ 
DiffuserCam~\cite{antipa2017diffusercam} & Hardware & 16.28\% & 4.26\% \\ 
DyPP~\cite{cheng2024learning} & Hardware & 39.52\% & 19.83\% \\ 
\midrule
\textbf{10-Layer LPS (Ours)} & \textbf{Hardware} & \textbf{38.38\%} & \textbf{18.36\%} \\ 
\textbf{LPS + MSPNet (Ours)} & \textbf{Hardware} & \textbf{67.90\%} & \textbf{9.05\%} \\ 
\bottomrule
\end{tabular}
}
\vspace{-1em}
\end{table}

\textbf{Analysis.} Software baselines (Gaussian Blur, Low Resolution) retain high identity recognition rates ($>$80\%), failing to provide effective privacy protection. Hardware methods (Coded Aperture, DiffuserCam) achieve low $ACC_S$ but severely degrade utility ($<$30\%) due to excessive scattering that destroys motion continuity. Our LPS combined with MSPNet achieves the best trade-off: suppressing $ACC_S$ to 9.05\% (strong privacy, $\sim$10$\times$ above random guess) while maintaining 67.90\% action recognition accuracy (only 11.62\% below the ideal level of 79.52\%)---a $\sim$28\% improvement over the next best hardware method (DyPP).

\subsubsection{Comparison with Clean Baseline}
To quantify the impact of LPS degradation, Table~\ref{tab:clean_baseline} compares performance on clean NTU RGB+D 120 versus privacy-protected P$^3$AR-NTU.

\begin{table}[t]
\centering
\caption{\textbf{Performance Drop from Clean to Privacy-Protected Data.} MSPNet exhibits the smallest degradation, demonstrating superior robustness to scattering.}
\label{tab:clean_baseline}
\renewcommand{\arraystretch}{1.2}
\resizebox{\columnwidth}{!}{
\begin{tabular}{l|c|c|c}
\toprule
\textbf{Method} & \textbf{Clean} & \textbf{P$^3$AR-NTU} & \textbf{Drop} \\
\midrule
SlowFast-ResNet50 & 88.52\% & 59.27\% & -29.25\% \\
VideoMAE-ViT-B & 86.37\% & 62.05\% & -24.32\% \\
InternVideo2-B/14 & 89.24\% & 65.97\% & -23.27\% \\
\midrule
\textbf{MSPNet (Ours)} & 79.52\% & \textbf{67.90\%} & \textbf{-11.62\%} \\
\bottomrule
\end{tabular}
}
\vspace{-4mm}
\end{table}

While large-scale pre-trained models achieve higher clean baselines due to massive pre-training on high-frequency visual details, they suffer severe degradation ($>$23\%) under LPS. In contrast, MSPNet---specifically designed for degraded videos---achieves the \textit{smallest performance drop} (11.62\%) and the \textit{highest accuracy} on the privacy-preserving task, confirming its robustness to scattering-induced degradation.

\subsection{Robustness to Reconstruction Attacks}
\label{sec:attack}
A critical concern for privacy-preserving imaging is whether the degradation can be inverted to recover private attributes. Following DyPP~\cite{cheng2024learning}, we define a hierarchical threat model with three levels of increasing attacker capability.

\textbf{Experimental Setup and Dataset.} Since the data-driven attack (Level C) requires paired Clean/LPS data to train reconstruction networks, we use paired data from the NTU RGB+D 120 subset (Setup ID $\geq$ S018) for training, while the test set uses \textbf{NTU RGB+D 60} (Setup ID S001--S017, 40 subjects) to avoid data leakage. All attack experiments in Table~\ref{tab:attack_results} are evaluated on this NTU60 subset. Note that the Clean baseline $ACC_S$ differs between this section (99.25\% on NTU60 with 40 subjects) and Section~\ref{sec:tradeoff} (98.36\% on NTU120 with 106 subjects) due to the different number of identity classes---random-guess accuracy is $\sim$2.5\% for 40 subjects vs.\ $\sim$0.94\% for 106 subjects. Despite this difference, both datasets consistently demonstrate that LPS effectively suppresses identity recognition far below their respective baselines.
\begin{itemize}[leftmargin=*]
    \item \textbf{Level A (Black-box):} The attacker only has access to LPS video streams and uses generic off-the-shelf deblurring tools (e.g., pre-trained Restormer). As shown in Table~\ref{tab:attack_results}, these methods actually \textit{degrade} image quality further ($ACC_S$ drops from 9.85\% to 7.21\%), as LPS degradation differs fundamentally from conventional blur.
    \item \textbf{Level B (PSF Inversion):} The attacker physically calibrates the effective PSF and performs non-blind deconvolution (Wiener, Richardson-Lucy).
    \item \textbf{Level C (Data-driven):} The attacker has paired Clean/LPS training data and trains a strong reconstruction network---the strongest but least realistic scenario.
\end{itemize}
We detail Level B and C below, as they represent more sophisticated attackers; Level A results are included in Table~\ref{tab:attack_results} for completeness.

\subsubsection{PSF Calibration and Characterization}
\begin{figure}[t]
    \centering
    \includegraphics[width=\linewidth]{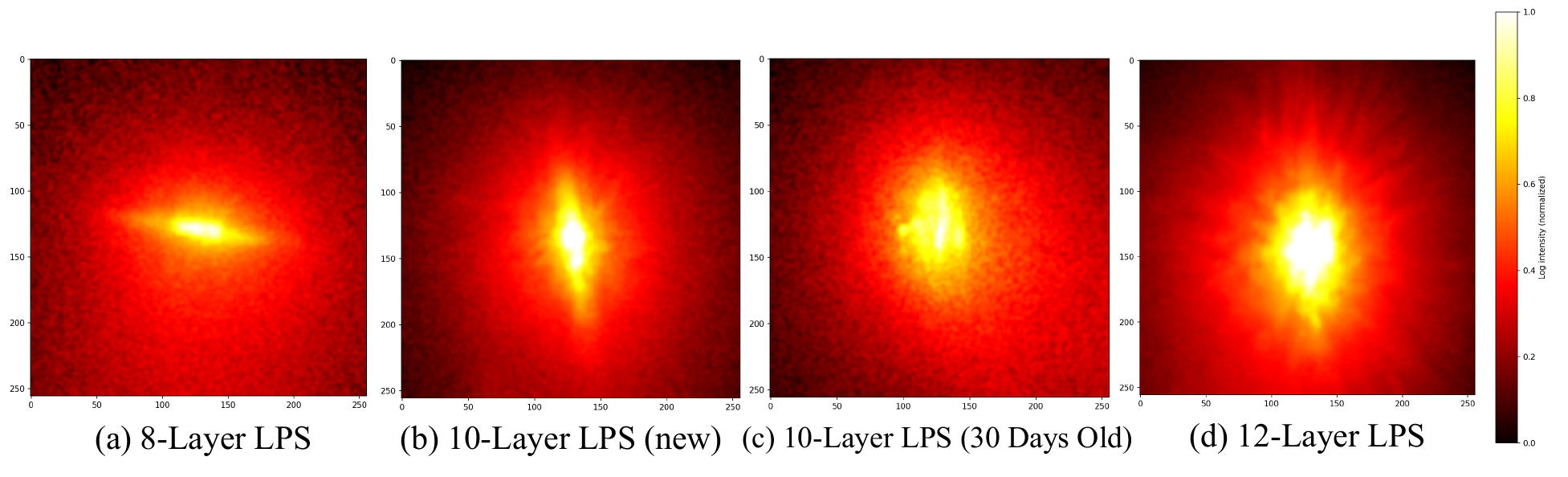}
    \vspace{-3mm}
    \caption{\textbf{Effective PSF of LPS under Different Configurations (Point Source Calibration, Log Scale).} From left to right: (a) 8-Layer LPS, (b) 10-Layer LPS (New), (c) 10-Layer LPS (30 Days Old), (d) 12-Layer LPS. The PSF exhibits significant non-Gaussian and stochastic scattering characteristics, changing dramatically with layer count and aging state. The 10-Layer PSF after 30 days shows obvious drift from the new state, demonstrating the PUF-like time-varying characteristics mentioned above.}
    \label{fig:psf_visualization}
    \vspace{-4mm}
\end{figure}

We first calibrated the effective Point Spread Function (PSF) of LPS through point-source imaging in a darkroom using a small LED source. PSFs were captured at the center and four corners to account for spatial variation, with multi-exposure HDR fusion to recover dynamic range. Fig.~\ref{fig:psf_visualization} visualizes the calibrated PSFs for different configurations.

The measured PSFs exhibit significant non-Gaussian and stochastic scattering characteristics: unlike regular circular spots from conventional defocus, LPS PSFs contain numerous irregular speckles and high-frequency noise caused by micro-bubbles, non-uniform adhesive layers, and surface roughness. This stochastic nature makes the inverse problem severely ill-posed.

\textbf{Optical Characterization Summary.} Our systematic experiments reveal: (1) \textit{MTF50} drops from 0.0729 to 0.0069 cycles/pixel with 10-layer LPS---a 90.5\% reduction in spatial resolution; (2) \textit{Veiling Glare Index} increases from 9.91\% to 71.04\%, indicating massive stray light contamination that drowns the signal; (3) \textit{Temporal drift}: 10-layer LPS exhibits 26.6\% transmittance change over 30 days, acting as a natural Physical Unclonable Function (PUF) that invalidates calibrated attack models over time; (4) \textit{Color distortion}: wavelength-dependent scattering causes significant spectral imbalance, further hindering attribute recovery.

\subsubsection{PSF Inversion Attack}
The attacker calibrates the effective PSF and performs non-blind deconvolution. We tested Wiener filtering (sweeping $K \in \{10^{-2}, 10^{-3}, 10^{-4}\}$) and Richardson-Lucy (iterations $\in \{5, 10, 20\}$) with the calibrated PSF.

As shown in Table~\ref{tab:attack_results} and Fig.~\ref{fig:psf_attack}, despite knowing the exact PSF, $ACC_S$ remains at 7.36\%---similar to direct LPS measurements. The restored images only retain coarse brightness structures while high-frequency identity features (facial details, textures) remain completely missing. This failure is explained by our optical characterization: $>$90\% high-frequency information loss makes the inverse problem fundamentally ill-posed.

\subsubsection{Data-driven Attack}
We trained Restormer~\cite{zamir2022restormer} on paired Clean/LPS data---the strongest attacker scenario. As described above, the training/test split follows the NTU60 protocol to avoid data leakage. Obtaining such paired data requires physically capturing the same scene simultaneously with and without LPS, which is practically infeasible in real deployments.

\begin{figure}[t]
    \centering
\includegraphics[width=\columnwidth]{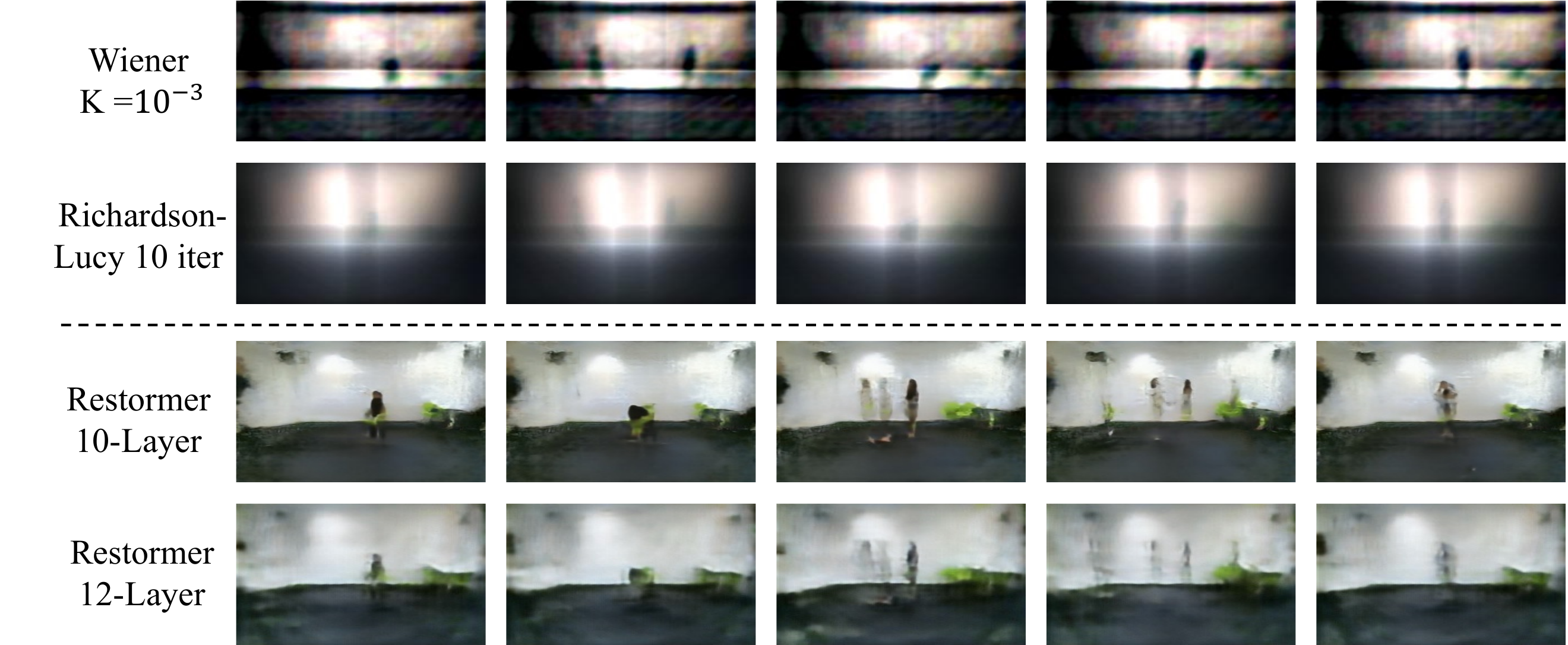}
    \vspace{-4mm}
    \caption{\textbf{Reconstruction Attack Visualization.} Calibrated PSFs show stochastic, non-Gaussian scattering. PSF inversion (Wiener, Richardson-Lucy) produces artifacts without recovering identity. Data-driven attack (Restormer) generates ``hallucinated'' generic faces, not actual identity features. The 12-Layer test shows degraded reconstruction quality, confirming cross-configuration generalization failure.}
    \label{fig:psf_attack}
    \vspace{-1em}
\end{figure}

As shown in Table~\ref{tab:attack_results} and Fig.~\ref{fig:psf_attack}, we evaluate the reconstruction attack on both 10-Layer (in-distribution) and 12-Layer (out-of-distribution) LPS. For 10-Layer test, while visual metrics improved (SSIM: 0.459$\to$0.533), $ACC_S$ only increased to 19.07\%. For 12-Layer test, the network trained on 10-Layer data fails to generalize: SSIM drops to 0.476 and $ACC_S$ remains at 14.72\%. This cross-configuration generalization failure is \textbf{favorable for privacy}: attackers cannot construct a universal reconstruction model to attack LPS devices with different layer configurations. Regardless of configuration, $ACC_S$ remains far below the clean baseline (99.25\%), and the ``recovered'' faces are statistically plausible hallucinations, not true identity features.

\textbf{Why Reconstruction Fails: Null Space and Cross-Configuration Analysis.}
The failure of reconstruction attacks confirms that the LPS forward operator $\mathcal{H}$ possesses a large \textit{null space} containing high-frequency biometric information. With VGI $>$70\% and MTF50 reduced by $>$90\%, identity-critical signals are irretrievably lost rather than merely attenuated, causing reconstruction networks to ``hallucinate'' generic face-like textures based on training priors rather than recovering actual identity features. The cross-configuration failure (10-Layer$\to$12-Layer) further reveals that optical characteristics change dramatically with layer count, preventing attackers from building universal reconstruction models. Despite SSIM improvement, running ArcFace~\cite{deng2019arcface} face verification on reconstructed vs.\ original faces yields similarity scores below 0.3 (random-level), confirming that the recovered textures are statistical artifacts, not identity-consistent features.

\begin{table}[t]
\centering
\caption{\textbf{Reconstruction Attack Results (NTU RGB+D 60).} Even with calibrated PSF or paired training data, identity recovery remains infeasible. The 12-Layer test shows that reconstruction networks trained on 10-Layer data fail to generalize across configurations.}
\label{tab:attack_results}
\renewcommand{\arraystretch}{1.2}
\resizebox{\columnwidth}{!}{
\begin{tabular}{l|cc|cc}
\toprule
\multirow{2}{*}{\textbf{Attack Method}} & \multicolumn{2}{c|}{\textbf{Image Quality}} & \multicolumn{2}{c}{\textbf{Task Performance}} \\
& SSIM $\uparrow$ & PSNR $\uparrow$ & $ACC_{act} \uparrow$ & $ACC_S \downarrow$ \\
\midrule
Clean (Baseline) & 1.000 & $\infty$ & 81.47 & 99.25 \\
\midrule
10-Layer LPS (Ours) & 0.459 & 26.45 & 69.81 & 9.85 \\
\midrule
\multicolumn{5}{l}{\textit{Level A: Black-box Attack}} \\
Wiener (Estimated Kernel) & 0.350 & 18.61 & 27.69 & 8.02 \\
Restormer (Pre-trained) & 0.374 & 21.95 & 20.69 & 7.21 \\
\midrule
\multicolumn{5}{l}{\textit{Level B: PSF Inversion Attack}} \\
Wiener (Calibrated PSF) & 0.435 & 23.86 & 38.74 & 9.36 \\
Richardson-Lucy (10 iter) & 0.472 & 25.51 & 34.03 & 9.67 \\
\midrule
\multicolumn{5}{l}{\textit{Level C: Data-driven Attack}} \\
Restormer (10-Layer Test) & 0.533 & 36.62 & 56.93 & 19.07 \\
Restormer (12-Layer Test) & 0.476 & 28.50 & 50.28 & 14.72 \\
\bottomrule
\end{tabular}
}
\vspace{-1em}
\end{table}

These experiments demonstrate that LPS provides \textit{physically irreversible} privacy protection under the evaluated threat models, achieved through stochastic multi-layer scattering that is fundamentally more robust than simple defocus or coded aperture systems.

\subsection{Robustness to Optical and Environmental Variations}
\label{sec:robustness}
A key concern for practical deployment is whether MSPNet can generalize across different optical configurations and challenging environmental conditions. We conduct comprehensive robustness experiments to address this.

\subsubsection{Cross-Configuration Generalization}
The optical properties of LPS vary significantly with the number of film layers and material aging. To evaluate whether a single model can operate robustly across different configurations, we train MSPNet on 10-Layer LPS (New) data and test on unseen configurations without fine-tuning.

\renewcommand{\arraystretch}{1.3}
\begin{table}[t]
\centering
\caption{\textbf{Cross-Configuration Generalization.} MSPNet trained on 10-Layer LPS generalizes well to unseen layer configurations and aged materials.}
\vspace{-3mm}
\begin{tabular}{l|c|c}
\toprule
\textbf{Testing Condition} & \textbf{$ACC_{act}\uparrow$} & \textbf{$ACC_S\downarrow$} \\
\midrule
10-Layer (New, In-distribution) & 67.90 & 9.05 \\
10-Layer (30 Days Old) & 66.71 & 8.62 \\
8-Layer & 69.25 & 17.93 \\
12-Layer & 62.83 & 7.92 \\
\bottomrule
\end{tabular}
\renewcommand{\arraystretch}{1}
\vspace{-4mm}
\label{tab:cross_config}
\end{table}

As shown in Table~\ref{tab:cross_config}, MSPNet demonstrates strong generalization: testing on 8-Layer achieves 69.25\% accuracy (even higher than 10-Layer due to lighter degradation), while 12-Layer yields 62.83\%. Notably, testing on 30-day-old 10-Layer LPS shows only a 1.27\% accuracy drop despite significant optical drift caused by material aging. This robustness stems from IFNS's frame-differencing mechanism, which cancels out static scattering patterns and focuses on motion dynamics that remain consistent across configurations.

\subsubsection{Environmental Robustness}
To validate performance under challenging real-world conditions, we designed five challenging environmental settings: (1) \textit{Low Light}: dusk with ambient light only ($<$50 lux); (2) \textit{Strong Backlight}: subject facing away from window with strong backlight; (3) \textit{Dynamic Background}: large TV at maximum brightness playing videos; (4) \textit{Camera Shake}: handheld capture with slight movement ($\sim$1m range); (5) \textit{Interfering Objects}: moving items in scene (waving objects, rolling balls). We also collected data under a \textit{Standard} condition (indoor normal lighting, static background, tripod-fixed camera) to augment training diversity. We recruited 9 new subjects (not in the original training set) to perform all 60 action categories under each condition, yielding $\sim$3,240 new clips. The training set includes data from all conditions, while the test set is split by subject with a 2:1 ratio.

\renewcommand{\arraystretch}{1.3}
\begin{table}[t]
\centering
\caption{\textbf{Environmental Robustness Results.} $\Delta$ indicates the accuracy change relative to the original P$^3$AR-PKU baseline (74.83\%).}
\vspace{-2mm}
\begin{tabular}{l|c|c}
\toprule
\textbf{Condition} & \textbf{$ACC_{act}$ (\%)} & \textbf{$\Delta$ (\%)} \\
\midrule
P$^3$AR-PKU (Reference) & 74.83 & --- \\
\midrule
Low Light & 71.26 & -3.57 \\
Strong Backlight & 69.70 & -5.13 \\
Dynamic Background & 67.99 & -6.84 \\
Camera Shake & 62.32 & -12.51 \\
Interfering Objects & 69.14 & -5.69 \\
\bottomrule
\end{tabular}
\renewcommand{\arraystretch}{1}
\vspace{-3mm}
\label{tab:env_robustness}
\end{table}

Table~\ref{tab:env_robustness} presents the results. MSPNet maintains robust performance under most challenging conditions: illumination variations (low light: -3.57\%, backlight: -5.13\%) are effectively handled by IFNS's frame differencing, which cancels out static lighting biases. Dynamic backgrounds (-6.84\%) and interfering objects (-5.69\%) cause moderate drops, as CFSA's semantic filtering and CLIP-based priors help distinguish human motion from background changes. The largest degradation occurs under camera shake (-12.51\%), which violates the static background assumption underlying IFNS. However, this scenario is rare in our target surveillance and smart-home applications where cameras are typically fixed.

\subsection{Ablation Study}
\label{sec:ablation}

\begin{table}[t]
\centering
\caption{Evaluation of IFNS \& CFSA}
\vspace{-2mm}
\renewcommand{\arraystretch}{1.3}
\begin{tabular}{r@{\hskip 5pt}c@{\hskip 5pt}c}
\toprule
\textbf{Method} & \textbf{$ACC_{act}\uparrow$} & \textbf{$ACC_S\downarrow$}\\
\hline
MSPNet & 67.90 & 9.05 \\
w/o CFSA & 50.96 & 8.83 \\
w/o IFNS \& CFSA & 38.38 & 18.36  \\
\bottomrule
\end{tabular}
\vspace{-4mm}
\renewcommand{\arraystretch}{1}
\label{tab:aba_new}
\end{table}

\subsubsection{Evaluation of IFNS \& CFSA}
\begin{figure*}[t]
    \centering
    \includegraphics[width=1\textwidth]{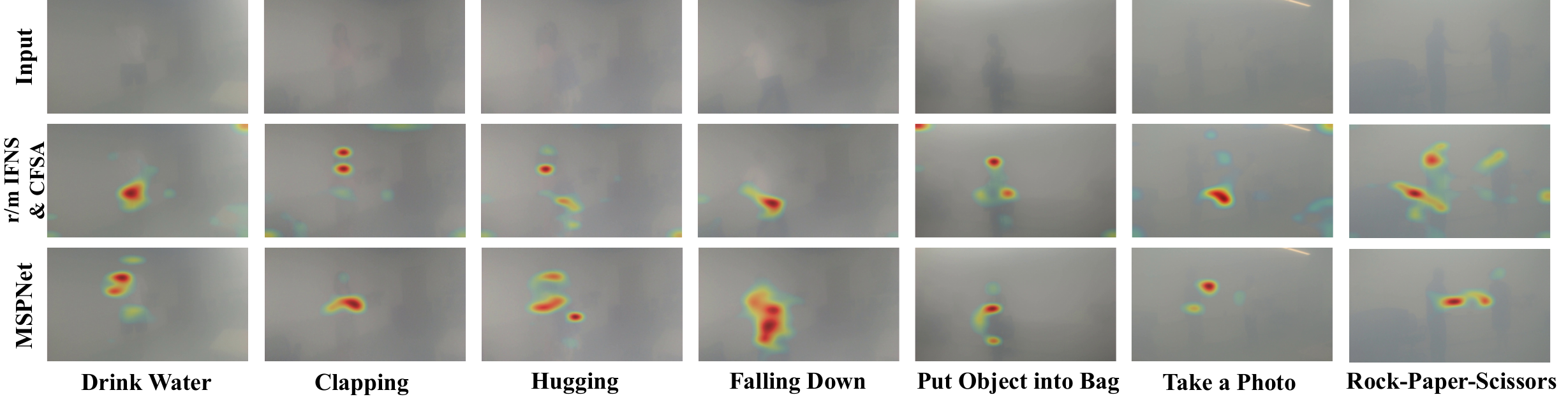}
    \vspace{-2mm}
    \caption{Attention Heatmap on P$^3$AR-PKU. The second row shows the results of MSPNet after removing IFNS and CFSA. We generate the attention heatmap by using the [CLS] token as the query vector and averaging the results across all attention heads to obtain the final visualization.}
    \vspace{-3mm}
    \label{fig:attention}
\end{figure*}

We evaluate the IFNS and CFSA modules in Table~\ref{tab:aba_new}, where removing both modules results in nearly 30\% decrease in action recognition accuracy and 9\% increase in $ACC_S$. Fig.~\ref{fig:attention} presents the attention heatmap visualization on the P$^3$AR-PKU dataset. Benefiting from the specifically designed IFNS and CFSA modules, the visualization results demonstrate enhanced focus on human motion-related regions. Removing CFSA independently leads to a 17\% accuracy decrease, confirming its effectiveness.

\subsubsection{Evaluation of cross-temporal frame differences in IFNS}
To validate the impact of cross-temporal frame differences in IFNS, we separately remove the residuals calculated with the previous frame, the subsequent frame, and the average frame. As shown in Table~\ref{tab:aba2_new}, each removal leads to accuracy decrease, with removing the average frame causing the most significant drop, underscoring its importance.

\begin{table}[t]
\centering
\caption{Evaluation of cross-temporal frame differences in IFNS}
\renewcommand{\arraystretch}{1.3}
\begin{tabular}{r@{\hskip 5pt}c@{\hskip 5pt}c}
\toprule
\textbf{Method} & \textbf{$ACC_{act}\uparrow$} & \textbf{$ACC_S\downarrow$}\\
\hline
MSPNet & 67.90 & 9.05 \\
w/o previous frame & 62.98 & 10.00  \\
w/o next frame & 61.50 & 9.70 \\
w/o average frame & 53.27 & 8.91  \\
\bottomrule
\end{tabular}
\vspace{-1em}
\renewcommand{\arraystretch}{1}
\label{tab:aba2_new}
\end{table}

\subsubsection{Evaluation of the number of steps in IFNS}
Table~\ref{tab:aba3_new} examines the impact of cross-temporal step count. Increasing steps up to four enhances spatio-temporal capture, but beyond four, excessive temporal disparity reduces useful information.

\begin{table}[t]
\centering
\caption{Evaluation of the number of steps in IFNS}
\renewcommand{\arraystretch}{1.3}
\begin{tabular}{r@{\hskip 5pt}c@{\hskip 5pt}c}
\toprule
\textbf{Method} & \textbf{$ACC_{act}\uparrow$} & \textbf{$ACC_S\downarrow$}\\
\hline
MSPNet ($T$ = $1$) & 63.95 & 8.73  \\
MSPNet ($T$ = $2$) & 66.89 & 8.98  \\
MSPNet ($T$ = $4$) & 67.90 & 9.05 \\
MSPNet ($T$ = $8$) & 66.58 & 9.52  \\
MSPNet ($T$ = $16$) &63.27& 10.27  \\
\bottomrule
\end{tabular}
\renewcommand{\arraystretch}{1}
\vspace{-1em}
\label{tab:aba3_new}
\end{table}

\subsubsection{Per-class Analysis and Limitations}
While MSPNet achieves strong overall performance, fine-grained actions involving subtle visual details (e.g., smiling, OK gesture) remain challenging due to physical blurring. As shown in Fig.~\ref{fig:pku60_app}(d), recognition accuracy for such actions is lower than for large-scale body movements. We address this from three perspectives: (1) \textit{Design Intent}: LPS prioritizes protecting biometric information; our target scenarios primarily require recognizing large-scale body movements. (2) \textit{Adjustable Granularity}: Users can customize the trade-off by adjusting film layers. (3) \textit{Motion Context Compensation}: For gesture-based actions, MSPNet captures arm motion trajectories through CFSA.

\section{Conclusion}
\label{sec:conclusion}
We introduce Lens Privacy Sealing (LPS), a hardware-level privacy solution using adjustable laminating films to provide pre-sensor protection at minimal cost. The P$^3$AR dataset benchmarks privacy-preserving action recognition with both large-scale and real-world subsets. MSPNet, incorporating IFNS and CFSA, achieves 67.90\% action accuracy while suppressing identity recognition to below 10\%.

Our comprehensive evaluation reveals three key findings: (1) LPS achieves a superior privacy-utility trade-off compared to state-of-the-art hardware methods (Coded Aperture, DiffuserCam, DyPP), maintaining significantly higher action accuracy at comparable privacy levels; (2) The stochastic multi-layer scattering of LPS provides physically irreversible privacy protection, resistant to both PSF inversion and data-driven reconstruction attacks; (3) MSPNet generalizes robustly across different optical configurations and challenging environmental conditions, with camera motion being the primary limitation.

Despite these advances, limitations remain. Fine-grained actions involving subtle visual details (e.g., micro-gestures) are challenging due to physical blurring. Camera motion significantly degrades performance as it violates the static background assumption. Additionally, our theoretical formulation models LPS as a linear operator, which is a first-order approximation; actual multi-layer scattering involves nonlinear effects that are not fully captured by this model. Nevertheless, the strong empirical performance of IFNS validates the practical utility of this approximation for action recognition. Future work will explore lightweight architectures for embedded deployment, adaptive film configurations for dynamic privacy levels, and extension to wearable and mobile scenarios.

\bibliographystyle{IEEEtran}
\bibliography{IEEEtran}
\begin{IEEEbiography}[{\includegraphics[width=1in,height=1.25in,clip,keepaspectratio]{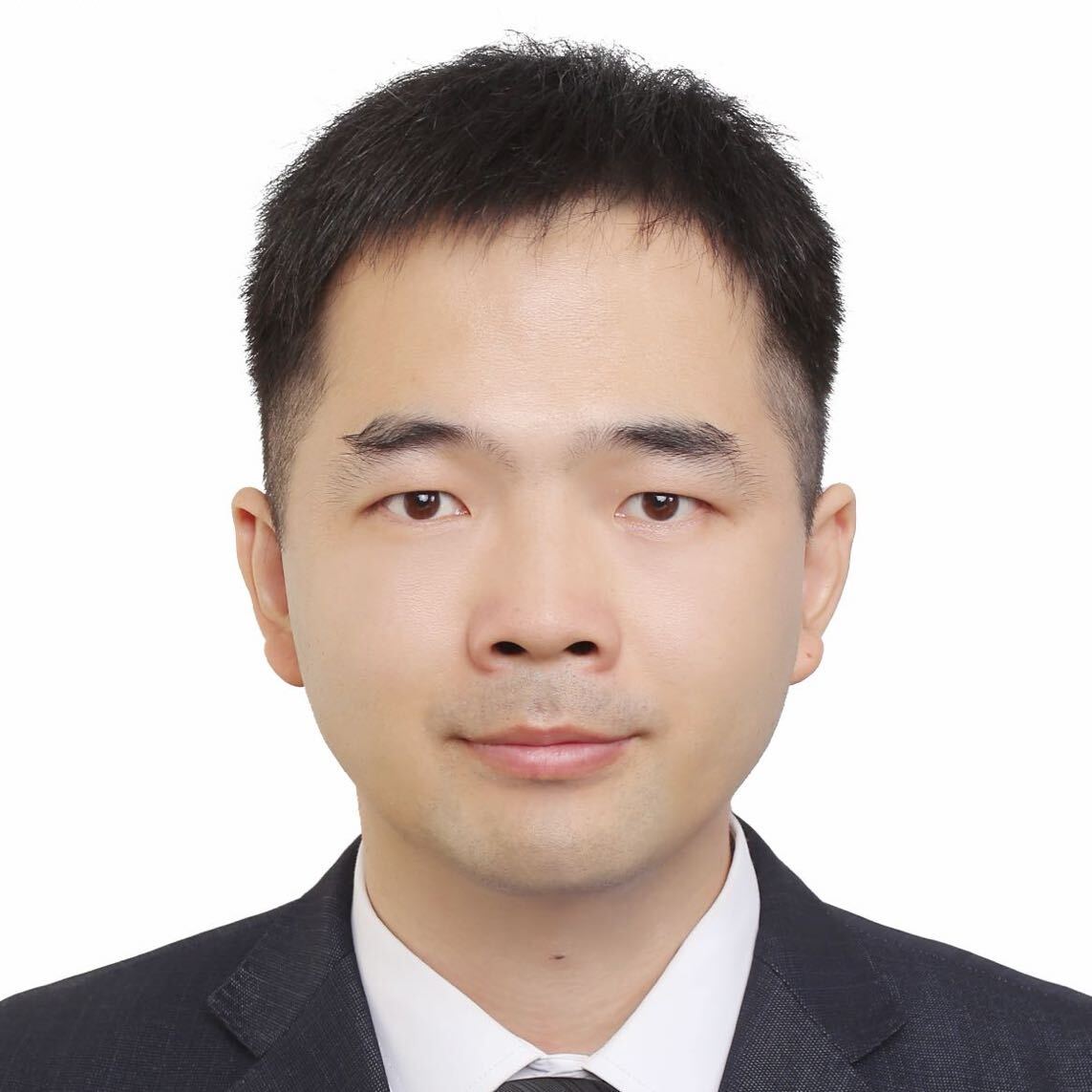}}]{Mengyuan Liu} received his Ph.D. under the supervision of Prof. Hong Liu at the School of Electrical Engineering and Computer Science (EE\&CS), Peking University (PKU), China. He subsequently worked as a research fellow with Prof. Junsong Yuan and Prof. Kai-Kuang Ma at the School of Electrical and Electronic Engineering, Nanyang Technological University (NTU), Singapore. Currently, he is an Assistant Professor and concurrently a Researcher at Peking University, Shenzhen Graduate School. His research primarily focuses on human-centric perception and human-robot interaction. His work has been published in leading conferences and journals, including NeurIPS, ICLR, ICML, CVPR, ICCV, TIP, TPAMI, and IJCV. In addition, he actively serves as an Associate Editor for Pattern Recognition (PR) and IEEE Transactions on Image Processing (TIP). \end{IEEEbiography}
\vspace{-2mm}
\begin{IEEEbiography}[{\includegraphics[width=1in,height=1.25in,clip,keepaspectratio]{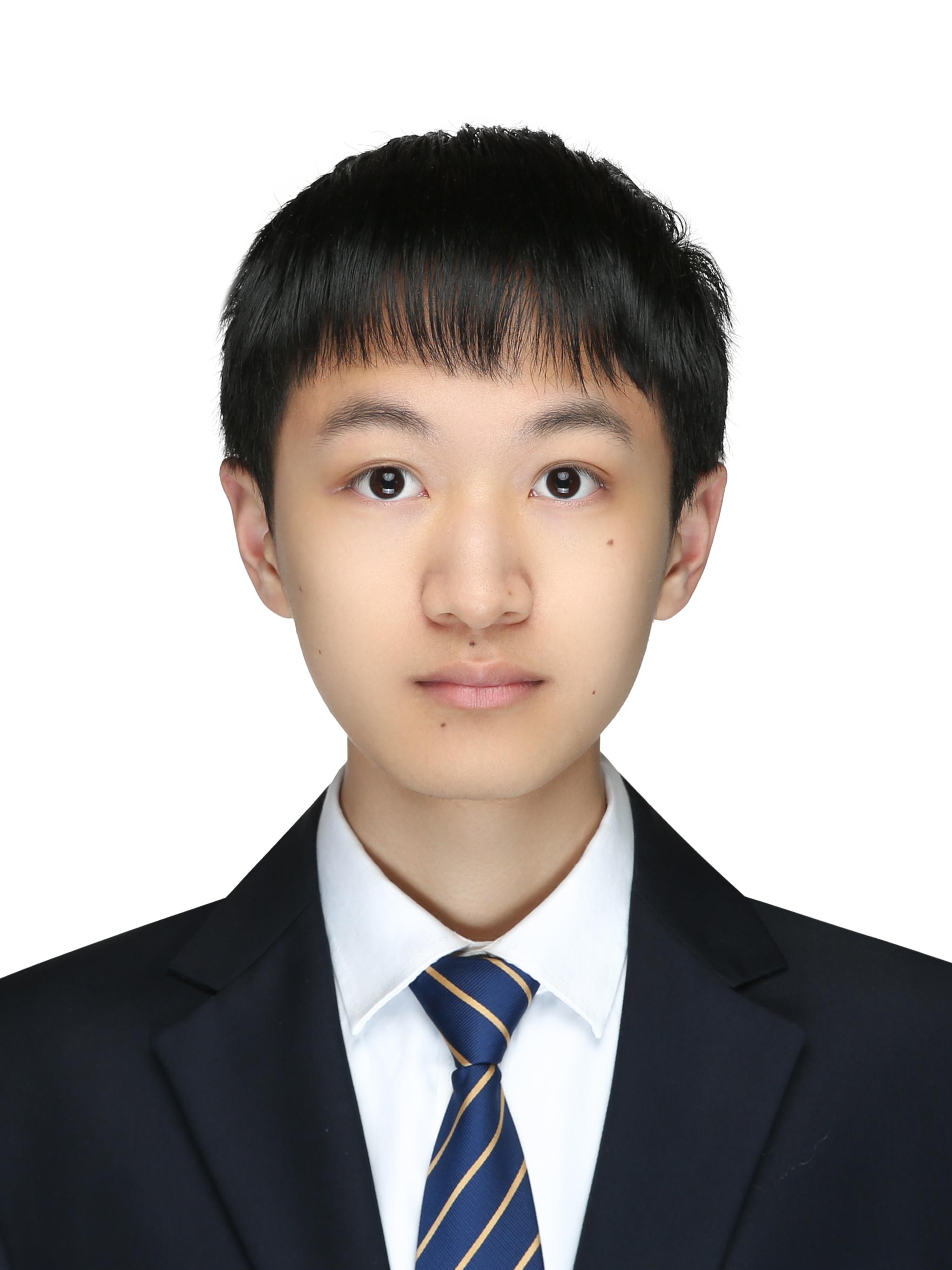}}]{Ziyi Wang} received the B.S. degree from Ocean University of China in 2024. He is currently pursuing the M.S. degree at Peking University Shenzhen Graduate School, advised by Asst. Prof. Mengyuan Liu. He has published papers in leading conferences, including ICML, ICCV, TIP, AAAI, and ACM MM. His research interests include human action recognition and multimodal large models. \end{IEEEbiography}
\begin{IEEEbiography}[{\includegraphics[width=1in,height=1.25in,clip,keepaspectratio]{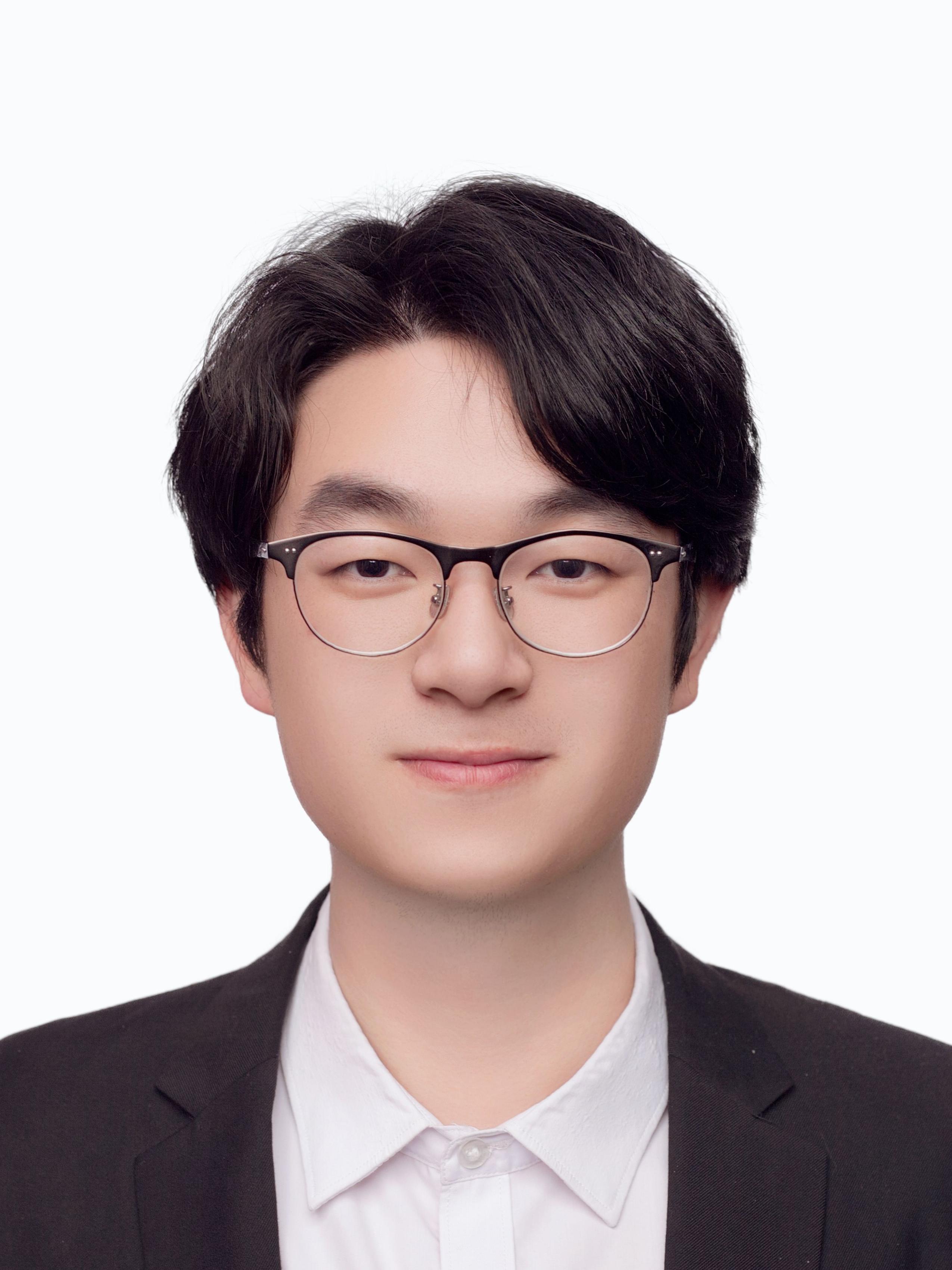}}]{Peiming Li} received the B.S. degree from Dalian University of Technology in 2024. He is currently pursuing the M.S. degree at Peking University Shenzhen Graduate School, advised by Asst. Prof. Mengyuan Liu. He has published papers in leading conferences, including ICML, ICCV, TIP, AAAI, and ACM MM. His research interests include human action recognition and multimodal large models. \end{IEEEbiography}
\begin{IEEEbiography}[{\includegraphics[width=1in,height=1.25in,clip,keepaspectratio]{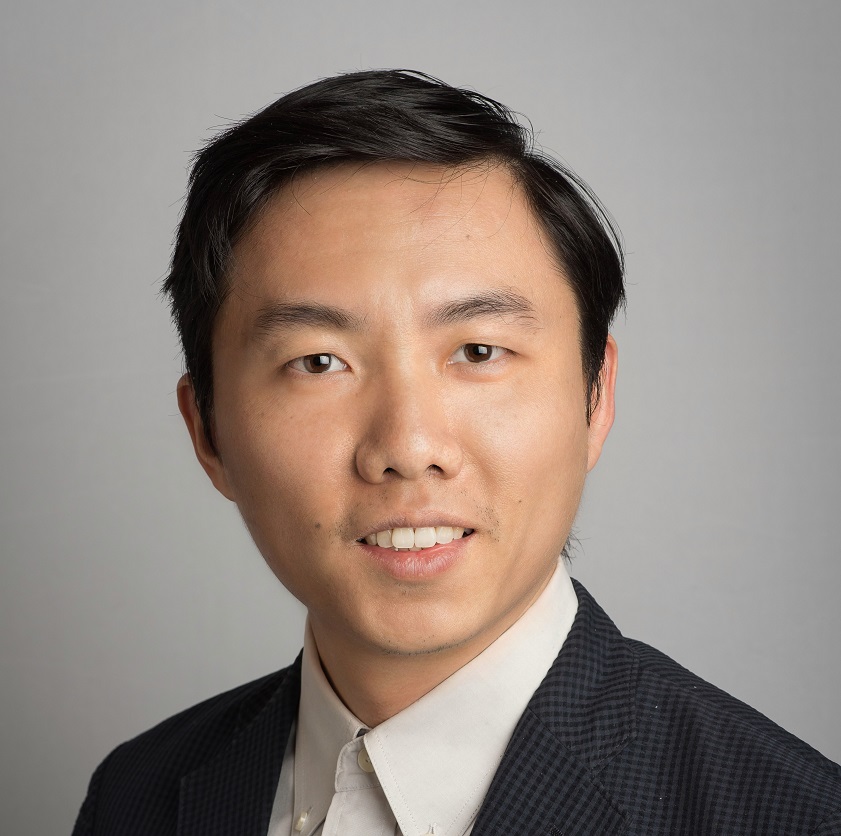}}]{Junsong Yuan} (Fellow, IEEE) received the B.Eng. degree from the Huazhong University of Science Technology (HUST) in 2002, the M.Eng. degree from the National University of Singapore in 2005, and the Ph.D. degree from Northwestern University in 2009. He is currently a Professor and the Director of the Visual Computing Laboratory, Department of Computer Science and Engineering (CSE), State University of New York at Buffalo, Buffalo, NY, USA. Before joining SUNY Buffalo, he was an Associate Professor (2015–2018) and Nanyang Assistant Professor (2009–2015) with Nanyang Technological University (NTU), Singapore. He is a fellow of IAPR.  He received the Chancellor’s Award for Excellence in Scholarship and Creative Activities from SUNY, Nanyang Assistant Professorship from NTU, the Outstanding EECS Ph.D. Thesis Award from Northwestern University, and the Best Paper Award from IEEE TRANSACTIONS ON MULTIMEDIA. He also serves as the General/Program Co-Chair for ICME and an Area Chair for CVPR, ICCV, ECCV, and ACM MM. He serves as a Senior Area Editor for \textit{Journal of Visual Communication and Image Representation} (JVCI) and an Associate Editor for IEEE TRANSACTIONS ON PATTERN ANALYSIS AND MACHINE INTELLIGENCE, IEEE TRANSACTIONS ON IMAGE PROCESSING, IEEE TRANSACTIONS ON CIRCUITS AND SYSTEMS FOR VIDEO TECHNOLOGY, and \textit{Machine Vision and Applications}. \end{IEEEbiography}

\end{document}